# Improved Semantic Segmentation of Tuberculosis-consistent findings in Chest X-rays Using Augmented Training of Modality-specific U-Net Models with Weak Localizations

**Sivaramakrishnan Rajaraman** [1,*], **Les Folio** [2], **Jane Dimperio** [2], **Philip Alderson** [3] and **Sameer Antani** [1]

1. Lister Hill National Center for Biomedical Communications, National Library of Medicine, Bethesda, Maryland 20814, United States of America; sameer.antani@nih.gov
2. Radiology and Imaging Sciences, Clinical Center, National Institutes of Health, Bethesda, Maryland 20814, United States of America; les.folio@nih.gov (L.F.); jane.dimperio@nih.gov (J.D.)
3. School of Medicine, Saint Louis University, St. Louis, Missouri 63103, United States of America; philip.alderson@health.slu.edu

* Correspondence: sivaramakrishnan.rajaraman@nih.gov; Tel.: +1-301-827-2383

**Abstract:** Deep learning (DL) has drawn tremendous attention in object localization and recognition for both natural and medical images. U-Net segmentation models have demonstrated superior performance compared to conventional hand-crafted feature-based methods. Medical image modality-specific DL models are better at transferring domain knowledge to a relevant target task than those that are pretrained on stock photography images. This helps improve model adaptation, generalization, and class-specific region of interest (ROI) localization. In this study, we train chest X-ray (CXR) modality-specific U-Nets and other state-of-the-art U-Net models for semantic segmentation of tuberculosis (TB)-consistent findings. Automated segmentation of such manifestations could help radiologists reduce errors and supplement decision-making while improving patient care and productivity. Our approach uses the publicly available TBX11K CXR dataset with weak TB annotations, typically provided as bounding boxes, to train a set of U-Net models. Next, we improve the results by augmenting the training data with weak localizations, post-processed into an ROI mask, from a DL classifier that is trained to classify CXRs as showing normal lungs or suspected TB manifestations. Test data are individually derived from the TBX11K CXR training distribution and other cross-institutional collections including the Shenzhen TB and Montgomery TB CXR datasets. We observe that our augmented training strategy helped the CXR modality-specific U-Net models achieve superior performance with test data derived from the TBX11K CXR training distribution as well as from cross-institutional collections ($p < 0.05$). We believe that this is the first study to i) use CXR modality-specific U-Nets for semantic segmentation of TB-consistent ROIs, and ii) evaluate the segmentation performance while augmenting the training data with weak TB-consistent localizations.

**Keywords:** deep learning; tuberculosis; convolutional neural networks; segmentation; modality-specific knowledge transfer; U-Net; chest-x-rays; augmentation; localization

## 1. Introduction

Tuberculosis (TB) is caused by mycobacterium tuberculosis and most often affects the lungs but can occur throughout the body. According to the 2020 World Health Organization (WHO) report [1], 10 million people were estimated to be infected with TB worldwide. While CT imaging is increasingly preferred, the simple chest X-ray (CXR) remains a widely used imaging modality in TB screening protocols, particularly in low and middle resource regions. Early diagnosis plays a crucial role in improving patient care and increasing the survival rate. However, there is a significant shortage of medical experts in the poor resource regions. Automated techniques have been proposed to offset some of these challenges. However, it is unclear if these methods are generalizable across different datasets or concur with radiologist findings. These concerns can be addressed using automated segmentation methods that determine a prediction model to obtain highly accurate results over the training dataset while maintaining good generalization for cross-institutional test sets. Next, segmentation results could be compared with expert radiologist assessments to determine performance. However, deep learning (DL) methods are known to be data-hungry in that reliable high-performing algorithms are dependent on good quality, large, varied, and well-labeled data. We are challenged by issues concerning varying image acquisition, the variability with which the TB-consistent findings may manifest in the images, and insufficiently or weakly labeled publicly available data sets. While we have little control over image quality, we can model the variation in disease presentation with richly labeled data. Here, we further discuss (i) the need for automated segmentation of TB-consistent findings, (ii) the need to train medical modality-specific segmentation models, and (iii) the methods proposed in this study to improve automated segmentation of TB-consistent findings in CXRs.

*1.1. Need for Automated Segmentation of TB-consistent Findings*

Manual expert TB segmentation and localization is a very challenging and time-consuming task in disease screening due to the sparse availability of experts, particularly in resource-limited countries across the world. Overcoming the lack of richly annotated and publicly available data is the impetus of our effort to create a generalizable DL method that can semantically segment TB-consistent findings on CXRs. We envision that such a method could help radiologists reduce errors following initial interpretation and before finalizing the report with "flags" (segmentations), indicating potential TB findings. This could improve radiologist accuracy and improve patient care through an additional "set of eyes" thereby augmenting the radiologist workflow to become more efficient. Automated segmentation could also serve as an initial point for radiology residents to better learn the characteristics of the disease manifestations by comparing their annotations of suspected abnormal findings and comparing with radiologist readings. Approved segmented findings could later be subjected to a multifactorial analysis including measuring the lobar location, shape, and other quantitative features confined to the ROI and interactive with workflows. This segmentation approach would lead to a significant advancement in ROI segmentation, localization, and diagnosis in resourced-constrained areas. To develop such a method, we need to localize, segment, and recognize TB-consistent findings on CXRs.

*1.2. Need to Train Medical Modality-specific Segmentation Models*

The challenge with TB is that the disease presents in a variety of shapes and sizes with different image intensities and may be obscured by or confused with superimposed or surrounding pulmonary structures. There are unique patterns of TB (e.g., cavities, nodule clusters) including anatomic distributions that support a TB etiology (e.g., apical predominance, para hilar, costophrenic angles). Medical image segmentation is often the first step in isolating anatomical organs, sites for dimension measurement, counting, disease classification, intervention planning, etc. It is a difficult task considering the existence of a high degree of variability in medical image quality to include penetration and positioning [2]. To overcome such variety, typical approaches in medical image segmentation develop solutions that are specific to the medical imaging modality, body part, or disease being studied [3]. This results in the limited direct use of segmentation models commonly trained on natural images for medical image segmentation [4]. There is an indispensable need to train generalizable medical modality-specific segmentation models to obtain high segmentation performance. The current literature leaves much room for progress in this regard. In this effort, we present our approach to this problem by building CXR modality-specific convolutional neural network (CNN) models for segmenting TB-consistent ROIs in CXRs. We provide background on these steps, i.e., classification and segmentation, in the following paragraphs.

We selected CNN models since they have demonstrated superior performance in natural and medical visual recognition tasks as compared to using conventional hand-crafted feature descriptors/classifiers [5]. CNNs are typically used in supervised learning-based classification tasks where the input image is assigned to one of the class labels. Recent advances reported in the literature use conventional hand-crafted feature descriptors/classifiers and DL models for classifying CXRs as showing normal lungs or pulmonary TB manifestations [6,7]. These studies reveal that the DL models outperform conventional methods toward medical image classification tasks, particularly CXR analysis [8]. The authors [7] used the publicly available TB datasets including the Shenzhen TB CXR, the Montgomery TB CXR [6], the Belarus TB CXR [9], and another private TB dataset from the Thomas Jefferson University Hospital, Philadelphia. The authors used ImageNet-pretrained AlexNet [10] and GoogLeNet [11] models to classify CXRs as showing normal lungs or pulmonary TB manifestations. To this end, the authors observed that an averaging ensemble of ImageNet-pretrained AlexNet and GoogLeNet models demonstrated statistically superior performance with an AUC of 0.99 ($p < 0.001$) compared to their untrained counterparts. Another study [12] used an atlas-based method to segment lungs and the scale-invariant feature transform (SIFT) algorithm to extract lung shape descriptor features from the CXRs in the Shenzhen TB CXR dataset to classify them as showing healthy or TB-infected lungs where they obtained an accuracy of 0.956 and an AUC of 0.99. In another study [13], the authors used Histogram of Oriented Gradient (HOG)-based feature descriptors with an SVM classifier to detect TB in CXRs in the India TB CXR collection [13], resulting in an accuracy of 0.942 and AUC of 0.957. In another study, a computer-aided diagnostic system was developed [14] using CNN models toward TB screening. The authors trained ImageNet-pretrained DL models on a private CXR data collection to classify the CXRs in the Shenzhen TB CXR collection as showing healthy or TB-infected lungs. The authors observed that the pretrained CNN models delivered a superior performance with an accuracy of 0.83 and AUC of 0.926 as compared to the baseline, untrained models. The authors [15] proposed a simple CNN model to classify the CXRs in the Shenzhen TB CXR, Montgomery TB CXR, and Belarus TB CXR collections as showing healthy or infected lungs. They observed that the custom CNN model achieved superior performance with an accuracy of 0.862 and an AUC of 0.925. Another study used ImageNet-pretrained CNN models to classify a private collection of 10,848 CXRs toward TB detection [16]. The authors obtained an accuracy of 0.903 and an AUC of 0.964.

Other studies demonstrate improvement in model robustness and generalization when the knowledge transfer is initiated from medical image modality-specific pretrained CNN models to a relevant medical visual recognition task [17]. Such a knowledge transfer, unlike conventional transfer learning, has demonstrated superior model adaptation to a relevant target task, particularly when the target task suffers from limited data availability. The authors propose the benefits offered through modality-specific pretraining on a large collection of CXR images, and then transferring and fine-tuning the learned knowledge toward detecting TB manifestations [18]. To this end, the authors observed state-of-the-art (SOTA) performance using a stacked model ensemble with the Shenzhen TB CXR collection with an accuracy of 0.941 and an AUC of 0.995. In another study [19], the authors proposed the benefits offered through modality-specific pretraining on a large collection of CXR images, and then transferring and fine-tuning the learned knowledge toward detecting TB manifestations. The authors observed superior performance with an accuracy of 0.9489 while constructing modality-specific model ensembles toward classifying CXRs as showing normal lungs or TB manifestations. However, there is no available literature that discusses modality-specific knowledge transfer applied to other visual recognition tasks like segmentation, particularly concerning CXR analysis.

For our segmentation task, the goal is to obtain region-based segmentation, i.e., each pixel in the input image has to be assigned to one of the class labels. However, the performance of these models is limited by the availability of annotated data. Medical imagery in particular is constrained by the highly limited availability of expert annotated samples. To overcome this limitation, a CNN-based model with a U-shaped architecture called the U-Net was proposed to analyze biomedical images [20]. The authors demonstrated that the U-Net model delivered superior pixel-level class predictions even under conditions of sparse data availability. The U-Net model has thus become the principal segmentation model architecture to be used for natural and medical image segmentation tasks. Several U-Net model variants have been proposed including V-Net [21], improved attention U-Net [22], nnU-Net [23], and U-Nets using ImageNet-pretrained encoders [24].

For CXR image analysis, U-Nets are prominently used to automate lung segmentation [25,26]. We find the literature is limited in other segmentation tasks, particularly for segmenting TB-consistent ROIs. There are no studies available at present that perform CXR-based TB-consistent ROI segmentation so that each pixel in the input image is assigned to one of the healthy or TB-consistent class labels. Thus, there is an essential need to (i) automate TB-consistent ROI segmentation through training SOTA segmentation models, and (ii) help clinicians localize these ROIs and supplement clinical decision-making.

*1.3. Proposed Methodology*

We propose a stage-wise methodology in this retrospective study.

i. We retrained a selection of ImageNet [27]-pretrained DL models including VGG-16 [28], VGG-19 [28], Inception-V3 [11], DenseNet-121 [29], ResNet-18 [30], MobileNet-V2 [31], EfficientNet-B0 [32], and NasNet-Mobile [33] on a large-scale collection of CXR images to convert the weight layers specific to the CXR modality and help learn CXR modality-specific feature representations.

ii. We propose CXR modality-specific VGG-16 and VGG-19 U-Nets hereafter referred to as the VGG-16-CXR-U-Net and VGG-19-CXR-U-Net models. These models use the CXR modality-specific pretrained VGG-16 and VGG-19 models from the previous step as the encoder backbone. The performance of these models and other SOTA U-Net model variants including the standard U-Net [20], V-Net with ResNet blocks [21], improved attention U-Net [22], ImageNet-pretrained VGG-16 U-Net [24], and ImageNet-pretrained VGG-19-U-Net are evaluated toward the lung segmentation task. The best performing model is used to segment lungs in a combined selection of CXRs showing normal lungs or pulmonary TB manifestations.

iii. We performed a knowledge transfer from CXR modality-specific pretrained models from (i) and fine-tuned them to classify the combined selection of CXRs from (ii) as showing normal lungs or pulmonary TB manifestations.

iv. The best performing fine-tuned model from (iii) is used to weakly localize the TB-specific ROI using class-selective relevance mapping (CRM) methods [34]. The localized ROIs are converted into binary ROI masks.

v. The segmentation models from (ii) are trained and evaluated to segment TB-consistent manifestations using CXRs. Here, we performed two sets of evaluations: (i) First, we used the publicly available, patient-specific train/test split of the TBX11K [35] dataset to train and evaluate the models; (ii) Next, we augmented the training data of the TBX11K dataset with the ROI masks generated from weak TB-consistent ROI localization from (iv) and their associated original CXRs from (ii) to further improve segmentation performance. This training process is hereafter referred to as augmented training (AT). The performance achieved with non-augmented training and AT using cross-institutional Shenzhen TB CXR [6] and Montgomery TB CXR [6] test collections is also evaluated to determine model robustness and generalization. Figure 1 illustrates the steps toward the current study.

We believe that this is the first study to i) use CXR modality-specific U-Net models to segment TB-consistent manifestations, and ii) evaluate non-augmented training and AT segmentation performance using test data derived from the same training data distribution and other cross-institutional collections to determine model robustness and generalization to real-time applications. The combined use of CXR modality-specific U-Net models and AT is expected to improve segmentation performance and could be applied to an extensive range of medical segmentation tasks including datasets from multiple institutions.

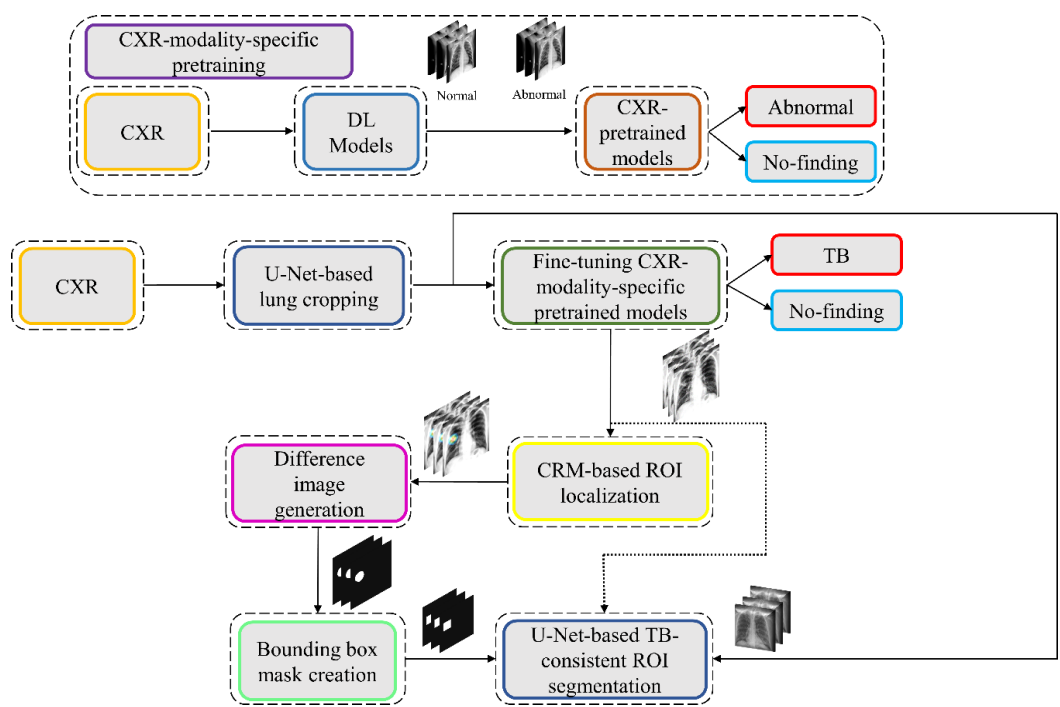

**Figure 1.** Graphical abstract of the proposed study.

The remainder of this study is organized as follows: Section 2 discusses the materials and methods, Section 3 elaborates on the results, Section 4 discusses the merits and limitations, and Section 5 concludes the study.

## 2. Materials and Methods

The materials and methods are further divided into the following sub-sections: (i) dataset characteristics; (ii) statistical analysis; (iii) CXR modality-specific pretraining; (iv) lung segmentation and preprocessing; (v) model fine-tuning and weak TB-consistent localization; (vi) verifying ROI localization; (vii) TB-consistent ROI segmentation using U-Net models; (viii) selecting appropriate loss function and other evaluation metrics; (ix) reducing inter-observer variability using Simultaneous Truth and Performance Level Estimation (STAPLE)-based consensus ROI generation, and (x) task-appropriate data augmentation.

*2.1. Dataset Characteristics*

The following publicly available CXR collections are used in this retrospective research study:

(i) Shenzhen TB CXR: This de-identified dataset contains 326 CXRs showing normal lungs and 336 abnormal CXRs showing various TB manifestations [6]. The CXRs are collected from Shenzhen No.3 hospital in Shenzhen, China. Each image is tested with diagnostic microbiology gold standard and acquired as a part of routine clinical care. It is exempted from institutional review board (IRB) review (OHSRP#5357) by the National Institutes of Health (NIH) Office of Human Research Protection Programs (OHSRP) and made publicly available by the National Library of Medicine (NLM). In this study, we split this dataset into two: (i) Shenzhen TB CXR-Subset-1 consists of 326 CXRs showing normal lungs and 268 CXRs showing pulmonary TB manifestations. This dataset is used to fine-tune the CXR modality-specific pretrained DL models to classify CXRs as showing normal lungs or TB manifestations. (ii) Shenzhen TB CXR-Subset-2 consists of 68 abnormal CXRs showing pulmonary TB manifestations. This dataset is used to perform cross-institutional testing toward TB-consistent ROI segmentation.

(ii) Montgomery TB CXR: This de-identified dataset is collected by the TB control program of the Department of Health and Human Services (HHS), Maryland, USA. The data set is exempted from institutional review board (IRB) review (OHSRP#5357) by the National Institutes of Health (NIH) Office of Human Research Protection Programs (OHSRP) and made publicly available by the NLM [6]. The collection includes 58 abnormal CXRs showing various TB manifestations and 80 CXRs showing normal lungs. Each image is tested with the diagnostic microbiology gold standard. Radiologist readings are made publicly available as text files. In this study, we used this dataset to perform cross-institutional testing toward TB-consistent ROI segmentation.

(iii) India TB CXR: The authors [13] from the National Institute of TB and respiratory diseases, New Delhi, India, released two different CXR datasets that are collected using different X-ray machines. Dataset A includes 78 CXRs for each class, showing normal lungs and others with various TB manifestations. These images were obtained with the Diagnox-4050 X-ray machine. Dataset B includes 75 CXRs for each class, showing normal lungs and other TB manifestations, and is acquired using the PRORAD URS X-ray machine with Canon detectors. This dataset is used in this study to fine-tune the CXR modality-specific pretrained DL models to classify CXRs as showing normal lungs or TB manifestations.

(iv) Belarus TB CXR: The International TB portals program at the National Institute of Allergic and Infectious Diseases (NIAID) is a leading scientific resource of annotated CXRs and CT images of TB patients [9]. The TB portal contains 298 CXRs showing various TB manifestations that are collected from patients in Belarus. This dataset is used to fine-tune the CXR modality-specific pretrained DL models to classify CXRs as showing normal lungs or TB manifestations.

(v) Tuberculosis X-ray (TBX11K) CXR: The authors [35] have made available a collection of 11,200 CXR images that are categorized into normal ($n$ = 5000), sick but not TB ($n$ = 5000), active TB ($n$ = 924), latent TB ($n$ = 212), active and latent TB ($n$ = 54) and uncertain TB ($n$ = 10) classes, where $n$ denotes the total number of CXRs in each class. Each image in the TBX11K collection is tested with diagnostic microbiology and annotated by the radiologists for TB manifestations. The images are de-identified and exempted for review by relevant institutions. The dataset also provides rectangular bounding box regions for TB positive cases. Note that using bounding boxes instead of fine annotation implicitly introduces errors in the training since a fraction of non-TB pixels will be treated as the positive class in pixel-wise training. This dataset is used to train and evaluate the U-Net segmentation models toward TB-consistent ROI segmentation.

(vi) Pediatric pneumonia CXR: This collection [36] includes 1583 anterior-posterior CXRs showing normal lungs and 4273 CXRs showing bacterial and viral pneumonia manifestations. The CXRs are collected from pediatrics of 1 to 5 years of age from the Guangzhou Women and Children's Medical Center, China. The images are acquired as a part of routine clinical care with IRB approvals. This dataset is used to perform CXR modality-specific pretraining of the ImageNet-pretrained CNN models toward classifying CXRs as showing normal lungs or other abnormal pulmonary manifestations.

(vii) Radiological Society of North America (RSNA) CXR: A part of the NIH CXR collection [37] is curated by the radiologists from RSNA and Society of Thoracic Radiology (STR) and made publicly available for a Kaggle challenge to detect pneumonia manifestations [38]. The collection includes 8851 CXRs showing normal lungs, 11,821 CXRs showing other abnormal manifestations but not pneumonia, and 6012 CXRs showing various pneumonia manifestations. The images are made available in DICOM format at 1024 × 1024-pixel resolutions. Ground truth (GT) disease annotations are made available for the CXRs belonging to the pneumonia class. This dataset is used in this study toward performing CXR modality-specific pretraining.

(viii) Indiana CXR: This collection [39] includes 2378 posterior-anterior CXRs showing abnormal pulmonary manifestations and 1726 CXRs showing normal lungs. These images are collected from various hospitals that are affiliated with the Indiana University School of Medicine. The dataset is de-identified, manually verified, archived at the NLM, and exempted from IRB review (OHSRP # 5357). This dataset is used during the CXR modality-specific pretraining stage. The datasets used in various stages of the proposed study and their distribution are summarized in Figure 2.

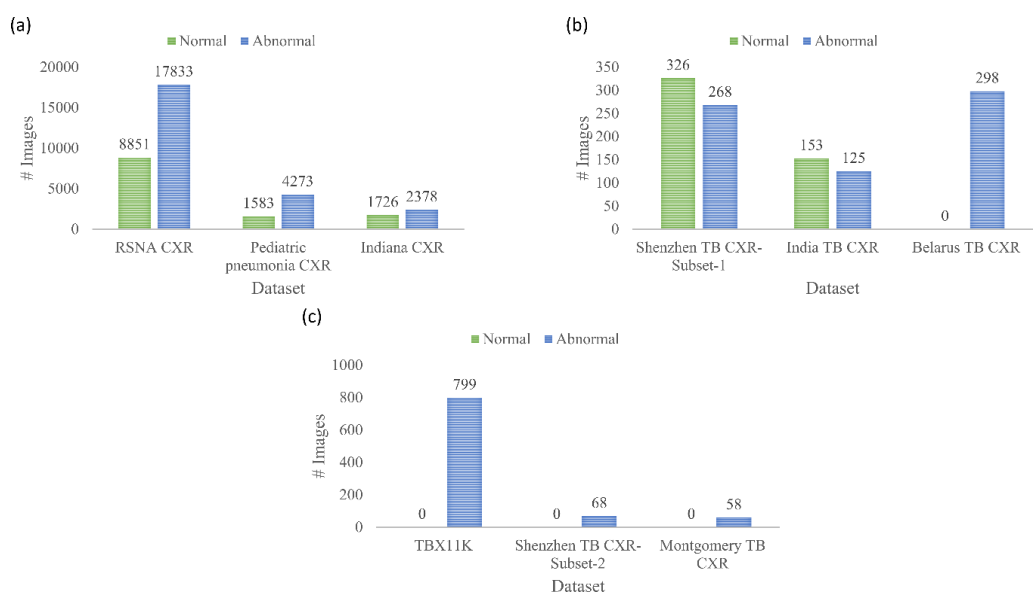

**Figure 2.** Datasets used in various stages of learning and their distribution. (**a**) CXR modality-specific pretraining; (**b**) fine-tuning; and (**c**) TB-consistent ROI segmentation.

## 2.2. Statistical Analysis

In this study, we used a 95% confidence interval (CI) to measure the error margin as the Binomial (Clopper-Pearson) "exact" method and discriminate models' performance. We used StatsModels v0.12.1 Python library to perform these analyses. Statistical approaches used only in subsets of this study will be described in those specific sections of the manuscript.

## 2.3. CXR Modality-specific Pretraining

In the first step of this stage-wise systematic study, we retrained a selection of ImageNet-pretrained DL models including i) VGG-16, ii) VGG-19, (iii) Inception-V3, (iv) DenseNet-121, (v) ResNet-18, (vi) MobileNet-V2, (vii) EfficientNet-B0, and (viii) NasNet-Mobile on a combined collection of RSNA CXR, pediatric pneumonia CXR, and Indiana CXR datasets to introduce sufficient data diversity in terms of image acquisition and patient demographics into the training process. The models are selected based on their architectural diversity and SOTA performance in other visual recognition tasks. Recall that the ImageNet-pretrained DL models are trained using stock photography images from the Internet. Using out-of-the-box ImageNet-pretrained models for medical visual recognition tasks may lead to model overfitting and loss of generalization. This is because the transfer of weights from the majority dataset does not adequately generalize on smaller medical datasets. Literature studies have shown that compared to using ImageNet-pretrained CNN models, CXR modality-specific model retraining (i) delivers superior performance toward classifying CXRs as showing normal lungs or other pulmonary abnormal manifestations and (ii) improve ROI localization performance with added benefits of reduced overfitting, prediction variance, and computational complexity [17]. CXR modality-specific pretraining converts the weight layers specific to the CXR modality and learns relevant features. The learned knowledge is transferred and fine-tuned to improve performance in a related target task. During this pretraining stage, the models are trained to classify CXRs as showing normal lungs or other abnormal manifestations.

The architectural depth and complexity of these models that are developed for natural visual recognition tasks may not be optimal for medical visual recognition because of issues concerning sparse data availability and varying feature representations. To this end, we performed empirical evaluations toward (i) identifying the best layer to truncate these models and (ii) appending task-specific heads to improve learning the underlying feature representation toward classifying the CXRs as showing normal lungs or other abnormal pulmonary manifestations. The truncated models are appended with the following task-specific layers: i) a 3 × 3 convolutional layer with 512 filters, (ii) a global average pooling (GAP) layer, (iii) a dropout layer (empirically determined ratio = 0.5), and (iv) a final dense layer with Softmax activation to output prediction probabilities for the normal and abnormal classes.

During this training step, the combined CXR collection including RSNA CXR, pediatric pneumonia CXR, and Indiana CXR datasets is split at the patient-level into 80% for training and 20% for testing. With a fixed seed value, we allocated 10% of the training data toward model validation. The models are optimized using stochastic gradient descent (SGD) algorithm to minimize the categorical cross-entropy loss toward this classification task. Callbacks are used to check model performance and the model checkpoints are stored after each epoch.

*2.4. Lung Segmentation and Preprocessing*

The performance of DL models is severely impacted by data quality. Irrelevant features may lead to biased learning and sub-optimal model performance. The task of detecting TB or other pulmonary disease manifestations is confined to the lung regions. Thus, regions in CXRs other than the lungs are irrelevant to be learned by the models. Hence, it is crucial to segment the lung regions and train the models on the lung ROI to help them learn relevant features concerning normal lungs or other pulmonary manifestations.

The U-Nets are one of the most powerful CNN models that are used for precise and accurate segmentation of medical images [20]. The principal advantage of U-Net is that it can handle data scarcity and learn from small training sets. The U-Net has a U-shaped architecture. It is composed of an encoder/contracting path and a decoder/expanding path and performs pixel-wise class segmentations. The feature maps from the various levels of the encoder are passed over to the decoder to predict features at varying scales and complexities.

In this study, we propose CXR modality-specific VGG-16 and VGG-19 U-Nets referred to as VGG-16-CXR-U-Net and VGG-19-CXR-U-Net models. These models use the CXR modality-specific pretrained VGG-16 and VGG-19 models from Section 2.3 as the encoder backbone. The architecture of the VGG-16-CXR-U-Net and VGG-19-CXR-U-Net models are shown in Figure 3. The encoder of the VGG-16-CXR-U-Net and VGG-19-CXR-U-Net models are each made up of five convolutional blocks, consisting of 13 and 16 convolutional layers, respectively, following the original VGG-16 and VGG-19 architecture. The green downward arrow at the bottom of each convolutional block in the encoder designates a max-pooling operation (2 × 2 filters, 2 strides) to reduce image dimensions. The decoder of the VGG-16-CXR-U-Net and VGG-19-CXR-U-Net models are each made up of five convolutional blocks and consist of 15 convolutional layers. The upward red arrow at the top of each convolutional block in the decoder signifies an up-sampling operation that performs transposed convolutions to restore the image to its original dimensions. Each convolutional layer in the encoder and decoder is followed by batch normalization (BN) and ReLU activation. The pink arrow signifies skip-connections that combine the corresponding feature maps and restores the original image dimensions. Sigmoidal activation is used at the final convolutional layer to predict binary pixel values.

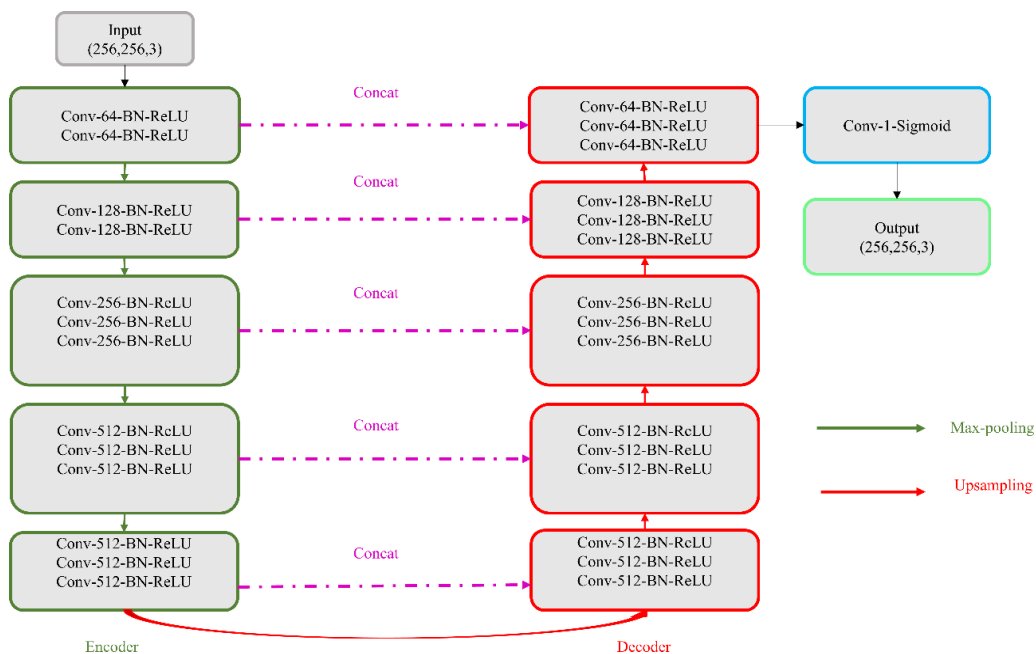

**Figure 3.** The architecture of the VGG-16- and VGG-19-CXR-U-Net models. The only difference is that the VGG-19-CXR-U-Net has an additional Conv-512-BN-ReLU layer in the convolutional blocks 3, 4, and 5 in the encoder path.

We evaluated the segmentation performance of the VGG-16-CXR-U-Net and VGG-19-CXR-U-Net models and other SOTA U-Net variants including the standard U-Net, V-Net with ResNet blocks, improved attention U-Net, ImageNet-pretrained VGG-16 U-Net, and VGG-19-U-Net toward lung segmentation. Figure 4 illustrates the lung segmentation workflow.

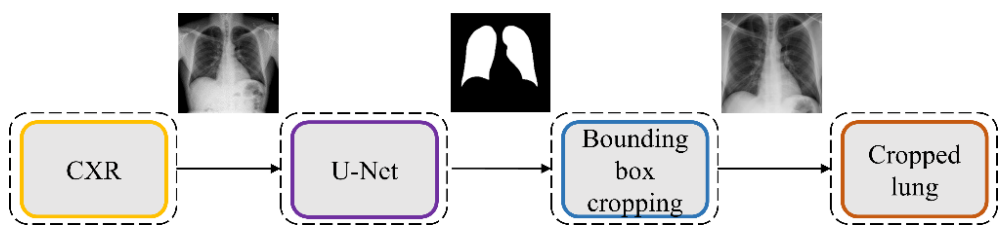

**Figure 4.** Lung Segmentation workflow. The best-performing U-Net model is used to generate lung masks that are overlaid on the original image to demarcate the lung boundaries and cropped to the size of a bounding box containing all lung pixels.

The models are trained on a publicly available collection of CXRs and their associated lung masks [40]. We allocated 10% of the training data with a fixed seed value for validation. We introduced variability into the training process by augmenting the training input with random affine transformations including horizontal flipping, height and width shifting, and rotations. The trained models are tested with the cross-institutional Montgomery TB CXR collection and their associated GT lung masks [6]. We emphasize that such an evaluation with a cross-institutional test collection would provide a faithful performance measure since the test data have varying visual characteristics and are completely unseen during model training, thereby preventing data leakage and ensuring generalization.

The best performing model is used to generate lung masks at 256 × 256 spatial resolution. The generated lung masks are overlaid on the original CXR images to demarcate the lung boundaries and then cropped into a bounding box encompassing the lung pixels. We pre-processed these images by (i) resizing the cropped bounding boxes to 256 × 256 spatial resolution, (ii) improving contrast by saturating the bottom and top 1% of all image pixel values, (iii) normalizing the pixel values to the range (0–1), and (iv) performing mean subtraction and standardization such that the resulting distribution has a mean μ of 0 and a standard deviation σ of 1 given by,

$$K_{norm} = \frac{K - \mu}{\sigma} \quad (1)$$

Here K is the original image and $K_{norm}$ is the normalized image.

### 2.5. Model fine-tuning and Weak TB-consistent Localization

The process workflow toward CXR modality-specific pretraining and fine-tuning is illustrated in Figure 5.

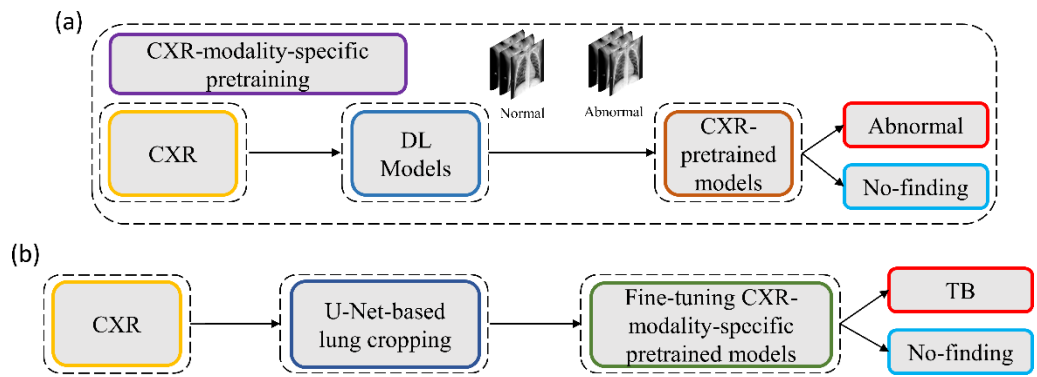

**Figure 5.** The process workflow showing (a) CXR modality-specific pretraining and (b) fine-tuning.

The CXR modality-specific pretrained models are truncated at their deepest convolutional layer and appended with the following task-specific layers: (i) GAP, (ii) dropout (empirically determined ratio = 0.5), and (iii) dense layer with two neurons. The resulting models are fine-tuned using the lung-segmented, combined TB CXR collection including India TB CXR, Belarus TB CXR, and Shenzhen TB CXR-Subset-1 datasets to classify them as showing normal lungs or pulmonary TB manifestations. The best-performing model from Section 2.4 is used to segment the lungs in this collection.

During this training step, the combined TB CXR collection is split at the patient-level into 80% for training and 20% for testing. We allocated 10% of the training data toward validation with a fixed seed value. The performance of these fine-tuned models is compared to their baseline counterparts, i.e., the out-of-the-box ImageNet-pretrained CNNs that are fine-tuned on this collection. The models are optimized using the SGD algorithm to minimize the categorical cross-entropy loss toward this classification task. We used callbacks to check the internal states of the fine-tuned models.

The following metrics are used to evaluate models' performance during CXR modality-specific pretraining and fine-tuning stages: (a) accuracy; (b) AUC; (c) sensitivity; (d) precision; (e) specificity; (f) F-measure; (g) Matthews correlation coefficient (MCC); and (h) diagnostic odds ratio (DOR).

### 2.6. Verifying ROI Localization

We used saliency maps and a CRM-based localization algorithm [34] to interpret the learned behavior of the fine-tuned models toward detecting TB manifestations. These algorithms provide a visual interpretation of model predictions and supplement clinical decision-making. The algorithms differ by the methods in backpropagating the derivatives and the use of feature maps. Saliency maps measure the derivative of the output category score concerning the original input and generate a heat-map with the original input image resolution. A higher derivative value signifies the importance of that activation in contributing to the final category score. A smaller value for the derivative results in negligible impact and the activation can thus be considered trivial toward final prediction.

CRM-based visualization algorithms are demonstrated to deliver superior localization performance as compared to the conventional class-activation map (CAM)-based localization, particularly toward medical image analysis [34]. Unlike CAM-based visualization, CRM-based localization underscores the fact that the feature maps contribute to decreasing the prediction scores for other class categories in addition to increasing the scores for the expected category. Such discrimination helps in maximizing the difference between these scores and results in superior discrimination of class-specific ROI by measuring the incremental mean-squared error from the output nodes. The features are extracted from the deepest convolutional layer of the best-performing fine-tuned model. The CRM algorithm is used to localize the ROI involved in classifying the CXRs as showing pulmonary TB manifestations. The feature map dimensions vary across the models. Hence, the CRMs are up-scaled through normalization methods to match the spatial resolution of the input image. The computed CRMs are overlaid on the original image to localize the TB-consistent ROI that is used to categorize the CXRs as showing pulmonary TB manifestations. The CRMs are generated for the CXRs with TB-category labels toward visualizing the regions of TB manifestations. We further converted these CRM-based weak TB-consistent ROI localizations to binary ROI masks. The original CXRs and the associated ROI masks are used for further analysis.

*2.7. TB-consistent ROI Segmentation Using U-Net Models*

The U-Net models used in this study are trained and evaluated on the publicly available train/test split of TBX11K CXR collection [35] to segment pulmonary TB manifestations. The dataset includes rectangular bounding box annotations for 599 training and 200 test samples. We used the best-performing model from Section 2.4 to segment the lungs in this collection. Following lung segmentation, we rescaled the disease bounding box coordinates and converted them into binary masks. Recall that since we are using bounding boxes for training a fraction of the pixels within these are false positive (FP) training labels which will adversely impact our outcome. These masks and their associated CXRs are used to train and test the models. We used a fixed seed value to allocate 10% of the training data toward validation. Variability is introduced into the training process by augmenting the training data through affine transformations including horizontal flipping, height and width shifting, and rotations. Callbacks are used to store model checkpoints and the best performing model is used to segment TB-consistent ROI.

We further evaluated the performance of the U-Net models with cross-institutional test sets. The Shenzhen TB CXR-Subset-2 ($n$ = 68) and the Montgomery TB CXR ($n$ = 58) collections are individually used as test sets toward this evaluation. Such cross-institutional testing would demonstrate the generalization ability of the models and their suitability for real-time deployment because the test set is diverse and unseen during the training process.

*2.8. Selecting Appropriate Loss Function and other Evaluation Metrics*

U-Net models, though having an excellent potential toward accurate medical image segmentation, often suffer from data imbalance. These issues are particularly prominent in applications that involve lung/TB-consistent ROI segmentation where the number of lung/TB-consistent ROI pixels is markedly lower compared to the total image pixels. Such imbalanced training may lead to learning bias and may adversely impact segmentation performance. To alleviate issues due to this data imbalance, a generalized loss metric based on the Tversky index has been proposed [22] that delivers superior performance compared to the conventional binary cross-entropy (BCE) loss. A superior trade-off between sensitivity and precision is demonstrated using the Tversky loss function for imbalanced data segmentation tasks. In this study, we customized the hyperparameters $\alpha$ and $\beta$ of the Tversky loss function given by,

$$TL(\alpha, \beta) = \frac{\sum_{k=1}^{K} a_{ok} b_{ok}}{\sum_{k=1}^{K} a_{ok} b_{ok} + \alpha \sum_{k=1}^{K} a_{ok} b_{1k} + \beta \sum_{k=1}^{K} a_{1k} b_{ok}} \quad (2)$$

Here a and b denote the set of predicted and GT binary labels, respectively, $a_{0k}$ is the probability of the pixel k to belong to the lung/TB-consistent ROI, $a_{1k}$ is the probability of the pixel k to belong to the background, $b_{0k}$ takes the value of 1 for a lung/TB-consistent ROI pixel and 0 for the background and vice versa for $b_{1k}$. Through empirical evaluations, we observed that the values for the hyperparameters $\alpha = 0.3$ and $\beta = 0.7$ demonstrated a good balance between precision and sensitivity and that higher values for $\beta$ resulted in improved performance and generalization while using imbalanced data and helped to boost sensitivity. Accordingly, we used these hyperparameter values in the current study. We used callbacks to store model checkpoints after the completion of an epoch. The best of the stored checkpoints is used as the final model for the subsequent analysis.

We measured segmentation performance in terms of the following metrics: (i) confusion matrix; (ii) Jaccard index, otherwise known as the intersection of union (IOU); (iii) Dice index; and (iv) average precision (AP). The IOU evaluation metric is widely used in image segmentation applications. It is given by a ratio as shown below:

$$IOU = \frac{TP}{TP + FP + FN} \quad (3)$$

Here TP, FP, and FN denote respectively true positives, false positives, and false negatives. Comparisons are made to GT from CXR interpretations by independent radiologists as described in Section 2.9. While comparing the predicted masks with the GT masks, a TP indicates that the predicted mask overlaps with the GT mask, exceeding a pre-defined IOU threshold. The abbreviation FP indicates that the predicted mask has no associated GT mask. FN indicates that a GT mask has no associated predicted mask. The Dice index is another evaluation metric widely used in segmentation and object detection tasks and is given by:

$$Dice = \frac{2TP}{2TP + FP + FN} \quad (4)$$

The Dice index is similar to and is positively correlated with IOU. Like the IOU, the value of the Dice index ranges from 0 to 1, with the latter signifying a higher similarity between the GT and predicted masks. We computed the AP as the area under the precision-recall curve (AUPRC) given by,

$$AP = \int_0^1 p(r) dr \quad (5)$$

Here *p* denotes precision and r denotes sensitivity/recall. The values of precision and recall are given by,

$$Precision = \frac{TP}{TP + FP} \quad (6)$$

$$Recall = \frac{TP}{TP + FN} \quad (7)$$

The value of AP lies within 0 and 1. In this study, we computed AP as an average over multiple IOU thresholds ranging from 0.5 to 0.95 (in increments of 0.05), denoted by AP@[.5:.95].

*2.9. Reducing Inter-observer Variability Using STAPLE-Based Consensus ROI Generation*

The GT disease annotations for Shenzhen TB CXR-Subset-2 are set by verification from two expert radiologists, hereafter referred to as R1 and R2. The GT annotations for the Montgomery TB CXR collection are set by the verification from two expert radiologists, hereafter referred to as R2 and R3. The expert R2 participated in both annotations. The collective experience of the experts counts to 65 years. The web-based, VGG Image Annotator tool [41] is used by radiologists to independently annotate these collections. The radiologists are asked to draw rectangular bounding boxes over the regions that they believed to show TB-consistent manifestations. We chose to use bounding boxes rather than fine segmentation to maintain similarity to the TBX11K data recognizing that there will be noisy pixels due to the bounding box as well as the variability from multiple expert annotators. The annotations are performed in independent sessions when the radiologists annotated the TB-consistent ROI in these CXR images. The annotations from individual radiologists are later exported to a JSON file for subsequent analyses. The experts' annotations would be made publicly available upon acceptance.

We used the STAPLE algorithm [42] to build a consensus ROI annotation from the experts' annotations for the Shenzhen TB CXR-Subset-2 and Montgomery TB CXR data collections. STAPLE is widely used for validating the segmentation performance of the models by comparing them to that of expert annotations. An expectation-maximization methodology is used where the probabilistic estimate of a reference segmentation is computed from a collection of expert annotations and weighed by each expert's estimated level of performance. The segmented regions are spatially distributed based on this knowledge, satisfying constraints of homogeneity. The steps involved in measuring the consensus ROI are as follows: Let $R = (r_1, r_2, ...., r_n)^N$ and $X = (x_1, x_2, ...., x_n)^N$ denote two column vectors, each containing *A* elements. The elements in *R* and *X* represent sensitivity and specificity parameters, respectively, characterizing one of *N* segmentations. Let B denote an $M \times N$ matrix that describes segmentation decisions made for each image pixel. Let *C* denote an indicator vector containing *M* elements representing hidden, true binary segmentation values. The complete data can be written as (*B*, *C*), and the probability mass function as $f(B, C | r, x)$. The performance level of the experts, characterized by a tuple (*r*, *x*) is estimated by the expectation-maximization algorithm, which maximizes (*r'*, *x'*), the data log-likelihood function, given by,

$$(r', x') = argmax_{r,x} \ln(f(B, C|r, x)) \quad (8)$$

The STAPLE-generated consensus ROI is then given by,

$$Consensus\ ROI = STAPLE(Masks_1, Masks_2) \quad (9)$$

Here $Masks_1, Masks_2$ denote the annotations of the experts that annotated the given CXR dataset. We used the STAPLE-generated consensus ROI as the standard reference, which reduces the aforementioned inter-reader variability across our multiple experts. Next, we converted these TB-consistent ROI coordinates into binary masks that are used as the GT masks to evaluate the aforementioned TB segmentation models. Figure 6 shows instances of TB-consistent ROI annotations made by the radiologists and the STAPLE-generated consensus ROI for a sample CXR instance from the Shenzhen TB CXR-Subset-2 and Montgomery TB CXR collections respectively.

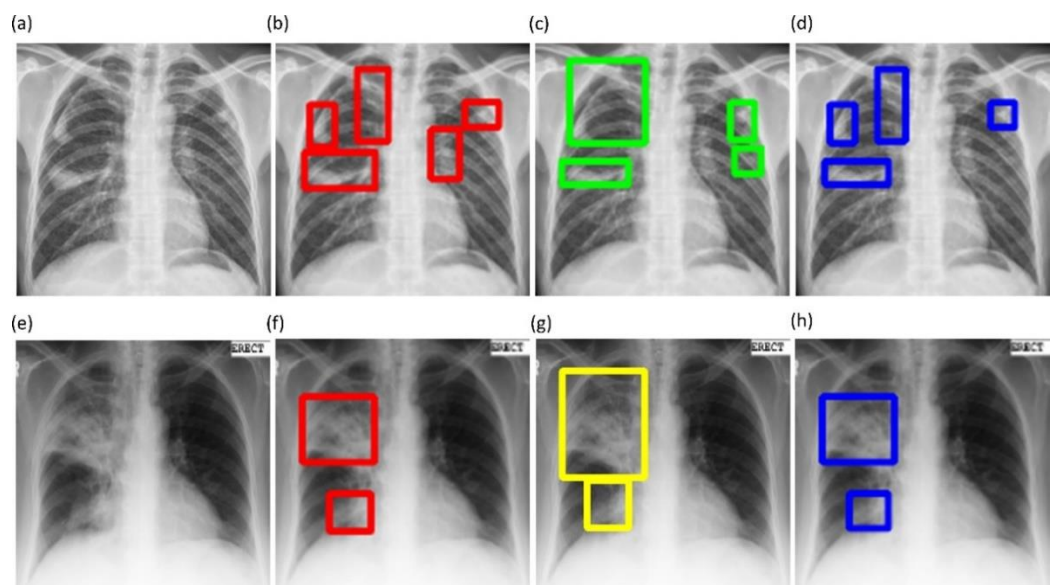

**Figure 6.** Examples showing annotations of the radiologists and the STAPLE-generated consensus ROI. (**a**) and (e) show the CXR image instances from the Shenzhen TB CXR-Subset-2 and Montgomery TB CXR datasets respectively; (**b**) and (**c**) show the annotations of R2 and R3 (bounding boxes in red and green), respectively for the given CXR instance from the Shenzhen TB CXR-Subset-2 dataset; (**f**) and (**g**) show the annotations of R2 and R1 (bounding boxes in red and yellow), respectively for the given CXR instance from the Montgomery TB CXR dataset; (**d**) and (**h**) show the STAPLE-generated consensus ROI (bounding boxes in blue) from the experts' annotations for the given CXR instance from the Shenzhen TB CXR-Subset-2 dataset and Montgomery TB CXR dataset respectively.

*2.10. Task-appropriate Data Augmentation: Augmenting Training Data with Weak Localizations to Improve TB-consistent ROI Segmentation*

The binary masks obtained from weak TB-consistent ROI localizations using the best-performing fine-tuned model and their associated CXR images are used to perform AT of the models used in Section 2.7 toward improving TB-consistent ROI segmentation. The performance with such AT is evaluated with the test set derived from TBX11K training distribution and individually with the cross-institutional Shenzhen TB CXR-Subset-2 and Montgomery TB CXR data collections. We used a fixed seed value toward allocating 10% of the training data toward validation. Variability is introduced into the training process by augmenting the training data through affine transformations including horizontal flipping, height and width shifting, and rotations. Callbacks are used to store model checkpoints and the best performing model is used to segment TB-consistent manifestations. The predicted masks are overlaid on the original CXR input to localize the TB-consistent ROI boundaries. The performance of the models with AT is evaluated and compared to those with non-augmented training using confusion matrix, IOU, Dice, and AP@[.5:.95] metrics. We used a Windows® system with NVIDIA GeForce GTX 1080 Ti GPU, Intel Xeon CPU E3-1275 v6 3.80 GHz processor, Keras framework with Tensorflow backend, and CUDA support for accelerated GPU performance toward these evaluations.

**3. Results**

The results are organized under the following sections: (i) CXR modality-specific pretraining; (ii) lung segmentation; (iii) fine-tuning toward TB detection; (iv) TB-consistent ROI localization and mask generation; (v) TB-consistent ROI segmentation evaluation using TBX11K train/test dataset, and (vi) TB-consistent ROI segmentation evaluation using TBX11K train set and cross-institutional test sets.

*3.1. CXR Modality-specific Pretraining*

Recall that during CXR modality-specific pretraining, we truncated the ImageNet-pretrained CNN models at the empirically determined intermediate layers that demonstrated superior performance toward this classification task. These layers are listed in Table 1. The naming conventions of these layers follow the Keras DL framework.

**Table 1.** Candidate layers that delivered superior performance during CXR modality-specific pretraining.

| Model | Truncated Layers |
|---|---|
| VGG-16 | Block5-conv3 |
| VGG-19 | Block5-conv4 |
| Inception-V3 | Mixed3 |
| DenseNet-121 | Pool3-pool |
| NASNet-mobile | Activation-94 |
| ResNet-18 | Add-6 |
| MobileNet-V2 | Block-9-add |
| EfficientNet-B0 | Block5c-add |

Table 2 shows the performance achieved by these models toward this classification stage.

**Table 2.** Performance measures achieved by the CXR modality-specific pretrained models.

| Model | ACC | AUC | Sens. | Spec. | Prec. | F | MCC | DOR |
|---|---|---|---|---|---|---|---|---|
| VGG-16 | 0.9009 | 0.9674 | 0.9009 | 0.8759 | 0.9026 | 0.9015 | 0.7799 (0.7664, 0.7934) | 64.1632 |
| VGG-19 | **0.9056** | **0.9687** | 0.9163 | **0.8841** | **0.9409** | **0.9284** | **0.7905 (0.7773, 0.8037)** | **83.5085** |
| Inception-V3 | 0.899 | 0.966 | 0.899 | 0.8759 | 0.901 | 0.8997 | 0.7761 (0.7626, 0.7896) | 62.8234 |
| DenseNet-121 | 0.8966 | 0.9602 | 0.8966 | 0.7912 | 0.896 | 0.895 | 0.7629 (0.7491, 0.7767) | 32.8575 |
| NasNet-Mobile | 0.8914 | 0.9609 | 0.8914 | 0.8224 | 0.8908 | 0.891 | 0.7535 (0.7395, 0.7675) | 38.0087 |
| ResNet-18 | 0.8881 | 0.9602 | 0.8881 | 0.8076 | 0.8872 | 0.8874 | 0.7451 (0.7309, 0.7593) | 33.3138 |
| MobileNet-V2 | 0.9015 | 0.9666 | 0.9015 | 0.8709 | 0.9027 | 0.9019 | 0.7803 (0.7668, 0.7938) | 61.7407 |
| EfficientNet-B0 | 0.905 | 0.9675 | **0.9196** | 0.8759 | 0.9372 | 0.9283 | 0.7882 (0.7749, 0.8015) | 80.7283 |

Data in parenthesis are 95% CI for the MCC values measured as the Binomial (Clopper-Pearson) "exact" method corresponding to separate 2-sided CI with individual coverage probabilities of √0.95. Acc. = Accuracy, AUC = Area under curve, Sens. = Sensitivity, Spec. = Specificity, Prec. = Precision, F = F-measure, MCC = Matthews correlation coefficient, DOR = Diagnostics odd ratio. The best performances are denoted by bold numerical values in the corresponding columns. None of these individual differences are statistically significant ($p > 0.05$).

It is observed from Table 2 that the VGG-19 model demonstrated superior performance for the accuracy, AUC, precision, specificity, F-measure, MCC, and DOR metrics. The DOR metric informs the discriminative power of the trained models. A higher value for DOR signifies high sensitivity and specificity with low FNs and FPs. The MCC metric offers more information than accuracy and F-measure because it considers a balanced ratio of TPs TNs, FPs, and FNs. The 95% CI for the MCC metric of the VGG-19 model demonstrated a tighter error margin and hence higher precision as compared to the other models. However, we noticed that these values are not statistically significantly different ($p > 0.05$) across the models. Considering the MCC and DOR values, the VGG-19 model demonstrated superior performance compared to other models. Figure 7 illustrates the confusion matrix, the area under receiver-operating-characteristic (AUC-ROC) curves, and the normalized Sankey diagram achieved by the VGG-19 model toward this classification task. The Sankey flow diagrams are used to illustrate energy flow and to assess the products' life-cycle [43]. In this study, we used Sankey flow diagrams to provide a visual representation of the models' performance. To this end, we assigned weights to the categories on the GT (left) and predictions (right) side to equally represent the categories on either side of the flow diagram. The width of the strips changes across the flow diagram such that the width of each strip at the right side illustrates the fraction of all images that the model predicts for a class that truly belongs to each class.

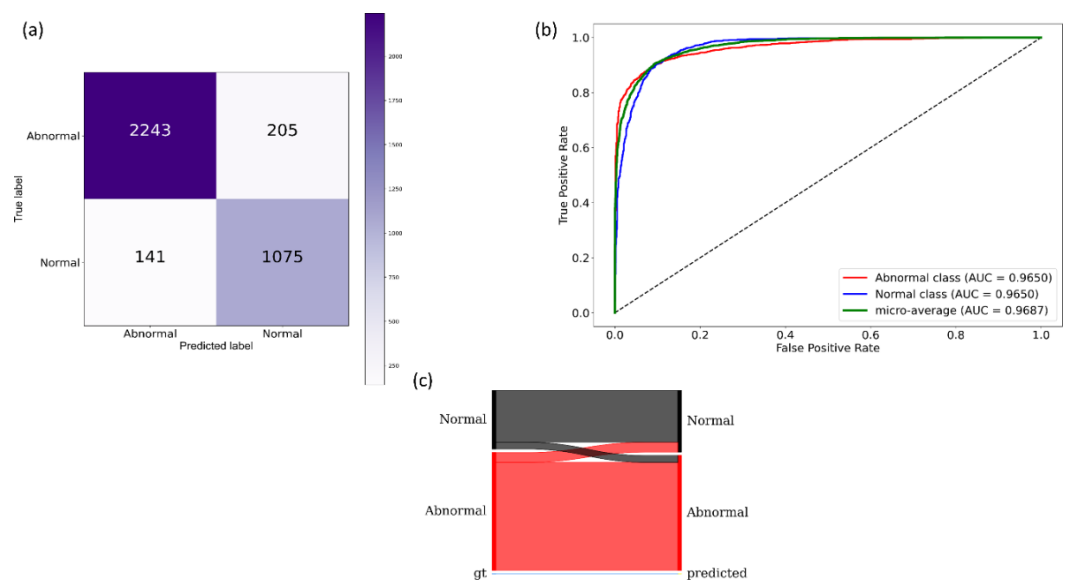

**Figure 7.** Performance achieved by the VGG-19 model during CXR modality-specific pretraining. (**a**) Confusion matrix; (**b**) AUC-ROC curves; (**c**) Normalized Sankey flow diagram.

*3.2. Lung Segmentation*

Table 3 shows the performance achieved by the SOTA U-Net variants and the proposed VGG-16-CXR-U-Net and VGG-19-CXR-U-Net models toward lung segmentation using the cross-institutional Montgomery TB CXR dataset for model testing.

**Table 3.** Performance measures achieved by the U-Net models toward CXR lung segmentation.

| Model | IOU | Dice | AP@[.5:.95] |
|---|---|---|---|
| Standard U-Net [5] | 0.9108 | 0.9529 | 0.8503 (0.7907,0.9099) |
| V-Net [6] | 0.9477 | 0.9729 | 0.9678 (0.9383,0.9973) |
| Improved attention U-Net [7] | 0.9493 | 0.9738 | 0.9548 (0.9201,0.9895) |
| VGG16-U-Net [8] | 0.9503 | 0.9744 | 0.9687 (0.9396,0.9978) |
| VGG19-U-Net | 0.9544 | 0.9765 | 0.9737 (0.947,1.000) |
| VGG16-CXR-U-Net (proposed) | 0.9532 | 0.9759 | 0.9715 (0.9437,0.9993) |
| VGG19-CXR-U-Net (proposed) | **0.9558** | **0.9774** | **0.9753 (0.9494,1.000)** |

Data in parenthesis are 95% CI for the AP@[.5:.95] values measured as the Binomial (Clopper-Pearson) "exact" method corresponding to separate 2-sided CI with individual coverage probabilities of $\sqrt{0.95}$. The best performances are denoted by bold numerical values in the corresponding columns. The performance of the standard U-Net is significantly worse compared to other models ($p < 0.05$). Otherwise, individual differences are not statistically significant.

It is observed from Table 3 that the proposed VGG19-CXR-U-Net demonstrated superior values for IOU, Dice, and AP@[.5:.95] metrics. The 95% CI for the AP@[.5:.95] metric obtained using the proposed VGG19-CXR-U-Net model demonstrated a smaller error margin and hence higher precision compared to other models. However, we noticed that these values are not statistically significantly different ($p > 0.05$) except for the standard U-Net model using which the performance achieved is statistically significantly different compared to other models ($p < 0.05$). Figure 8 illustrates the AUC-ROC curve and confusion matrix obtained using the proposed VGG19-CXR-U-Net toward the lung segmentation task.

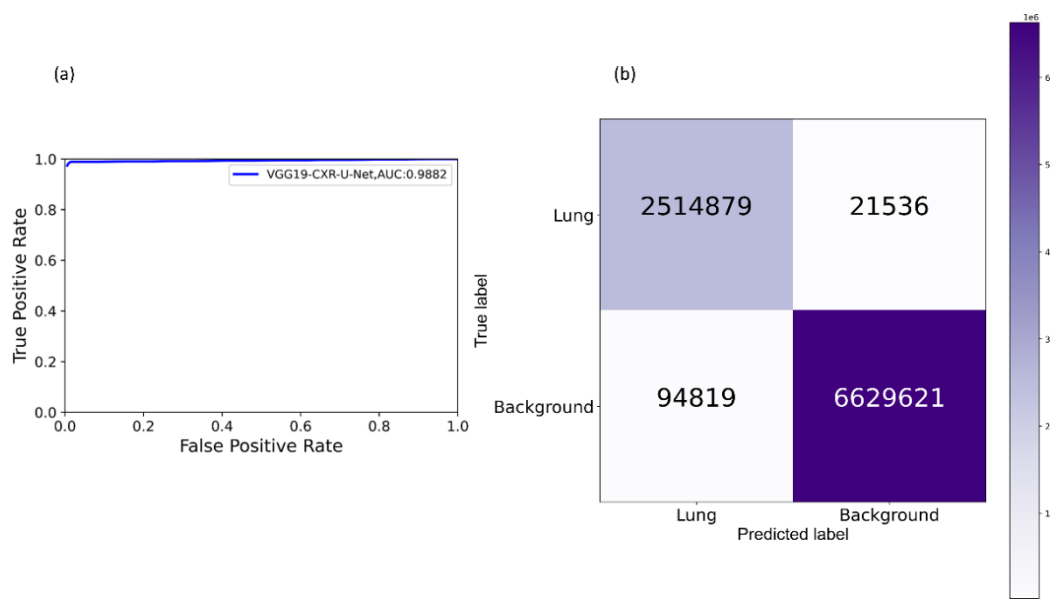

**Figure 8.** Performance achieved by the proposed VGG19-CXR-U-Net model toward lung segmentation task. (**a**) AUC-ROC curve; (**b**) Confusion matrix.

It is observed from the confusion matrix that the model demonstrates lower FNs compared to FPs and hence higher sensitivity compared to precision. That is, fewer lung pixels are classified as belonging to the background. This is because we empirically determined the hyperparameter values ($\alpha = 0.3$ and $\beta = 0.7$) of the Tversky loss function so that higher values for $\beta$ resulted in improved performance and helped to boost sensitivity. Figure 9 shows an instance of a CXR image on which the generated lung mask is overlaid to delineate the lung boundaries.

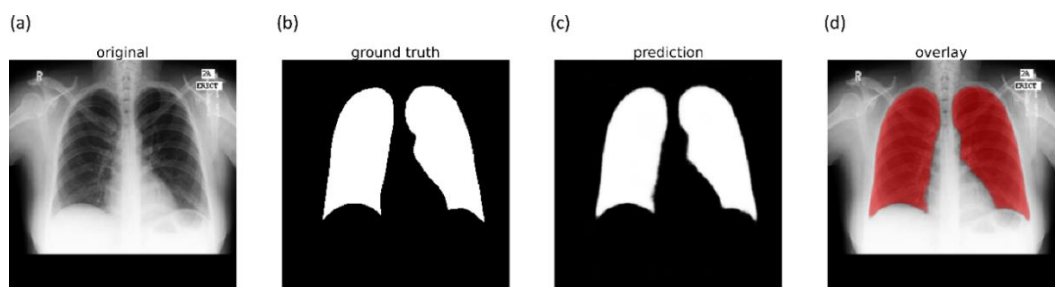

**Figure 9.** Sample lung segmentation result. (**a**) An instance of Montgomery TB CXR; (**b**) GT mask; (**c**) Predicted mask; and (**d**) Predicted mask overlaid on the original image.

*3.3. Fine-tuning Toward TB Detection*

During fine-tuning, the CXR modality-specific pretrained models are truncated at their deepest convolutional layer and appended with a task-specific head to classify CXRs as showing normal lungs or pulmonary TB manifestations. The performance achieved by the fine-tuned models is compared to their baseline counterparts, i.e., out-of-the-box ImageNet-pretrained models that are fine-tuned toward this classification task and are shown in Table 4.

**Table 4.** Performance metrics achieved by the fine-tuned models and their baseline counterparts.

| Models | Acc. | AUC | Sens. | Spec. | Prec. | F | MCC | DOR | Redux. |
|--------|------|-----|-------|-------|-------|---|-----|-----|--------|

| Model | Acc | AUC | Sens | Spec | Prec | F | MCC (95% CI) | DOR | Redux. |
|---|---|---|---|---|---|---|---|---|---|
| VGG-16-Baseline | 0.8828 | 0.9434 | 0.8766 | 0.8919 | 0.9221 | 0.8988 | 0.7612 (0.7106, 0.8118) | 58.575 | |
| VGG-16-Finetuned | **0.9231** | 0.9532 | 0.9692 | 0.8559 | 0.9076 | **0.9374** | **0.8411 (0.7977, 0.8845)** | **186.4375** | 0 |
| VGG-19-Baseline | 0.9011 | 0.9441 | 0.9383 | 0.8469 | 0.8995 | 0.9185 | 0.7942 (0.7462, 0.8422) | 84.0471 | |
| VGG-19-Finetuned | 0.9158 | **0.963** | 0.9198 | **0.9100** | **0.9372** | 0.9284 | 0.8264 (0.7814, 0.8714) | 115.7616 | 0 |
| Inception-V3-Baseline | 0.8754 | 0.9304 | 0.9198 | 0.8109 | 0.8765 | 0.8976 | 0.7404 (0.6883, 0.7925) | 49.1209 | |
| Inception-V3-Finetuned | 0.9048 | 0.9456 | 0.9198 | 0.8829 | 0.9198 | 0.9198 | 0.8027 (0.7554, 0.8500) | 86.4024 | 42.82 |
| DenseNet-121-Baseline | 0.8645 | 0.9288 | 0.8519 | 0.8829 | 0.914 | 0.8818 | 0.7260 (0.6730, 0.7790) | 43.3462 | |
| DenseNet-121-Finetuned | 0.8974 | 0.9399 | 0.9383 | 0.8379 | 0.8942 | 0.9157 | 0.7866 (0.7379, 0.8353) | 78.5334 | **56.52** |
| NasNet-Mobile-Baseline | 0.8828 | 0.9403 | 0.9013 | 0.8559 | 0.9013 | 0.9013 | 0.7571 (0.7062, 0.8080) | 54.1797 | |
| NasNet-Mobile-Finetuned | 0.8865 | 0.9258 | 0.9692 | 0.7658 | 0.858 | 0.9102 | 0.7679 (0.7178, 0.8180) | 102.6539 | 11.64 |
| ResNet-18-Baseline | 0.8865 | 0.9371 | 0.9075 | 0.8559 | 0.9019 | 0.9047 | 0.7644 (0.7140, 0.8148) | 58.1875 | |
| ResNet-18-Finetuned | 0.9048 | 0.9416 | 0.9013 | **0.9100** | 0.9359 | 0.9183 | 0.8052 (0.7582, 0.8522) | 92.1625 | 44.56 |
| MobileNet-V2-Baseline | 0.8645 | 0.9188 | 0.8766 | 0.8469 | 0.8931 | 0.8848 | 0.7206 (0.6673, 0.7739) | 39.2589 | |
| MobileNet-V2-Finetuned | 0.8865 | 0.9242 | **0.9754** | 0.7568 | 0.8541 | 0.9107 | 0.7694 (0.7194, 0.8194) | 122.8889 | 36.35 |
| EfficientNet-B0-Baseline | 0.9194 | 0.9548 | 0.9383 | 0.8919 | 0.9269 | 0.9326 | 0.8327 (0.7884, 0.8870) | 125.4 | |
| EfficientNet-B0-Finetuned | 0.9194 | 0.9442 | 0.963 | 0.8559 | 0.907 | 0.9342 | 0.8331 (0.7888, 0.8774) | 154.375 | 44.17 |

Data in parenthesis are 95% CI for the MCC values measured as the Binomial (Clopper-Pearson) "exact" method corresponding to separate 2-sided CI with individual coverage probabilities of √0.95. Redux. = Reduction in trainable parameters in %. The Baseline signifies fine-tuning out-of-the-box ImageNet-pretrained CNNs toward this classification task. The best performances are denoted by bold numerical values in the corresponding columns. Except for the NasNet-Mobile and EfficientNet-B0 models, all other fine-tuned models demonstrated statistically significantly superior performance ($p < 0.05$) for their MCC metric compared to their baseline counterparts.

As shown in Table 4, the models that were CXR-specific fine-tuned achieved superior performance compared to their ImageNet-trained baseline counterparts. Except for the NasNet-Mobile and EfficientNet-B0 models, all other fine-tuned models demonstrated statistically significantly superior performance ($p < 0.05$) for their MCC metric compared to their baseline counterparts. The VGG-16 fine-tuned model demonstrated superior values for accuracy, F-measure, MCC, and DOR metrics. The MCC value for this model demonstrated a smaller error margin and hence higher precision compared to other models. A significant reduction in the computational parameters is demonstrated by the fine-tuned models as compared to their baseline counterparts. The fine-tuned DenseNet-121 demonstrated a 56.52% reduction in the trainable parameters while delivering superior performance compared to the out-of-the-box ImageNet-pretrained counterpart. The same holds valid with other fine-tuned models except for VGG-16 and VGG-19 models where we had a similar number of trainable parameters compared to baseline models. Figure 10 shows the confusion matrix, AUC-ROC curves, and the normalized Sankey diagram achieved by the VGG-16 fine-tuned model toward this classification task.

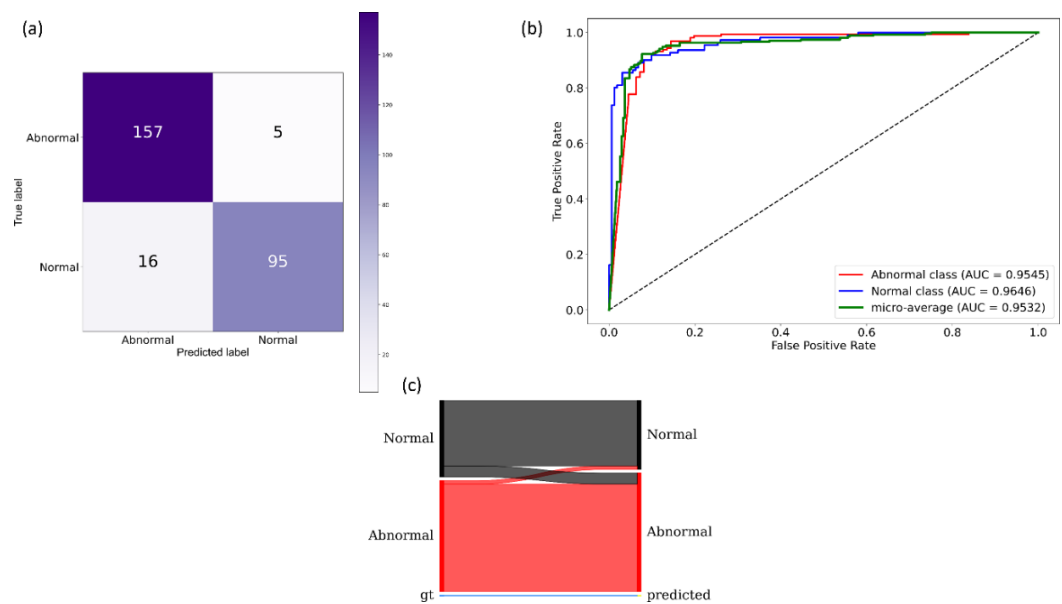

**Figure 10.** Performance achieved by the VGG-16 fine-tuned model. (**a**) Confusion matrix; (**b**) AUC-ROC curves; (**c**) Normalized Sankey flow diagram.

*3.4. TB-consistent ROI Localization and Mask Generation*

We studied the saliency maps to interpret the learned behavior of the VGG-16 fine-tuned model that delivered superior performance in classifying CXRs as showing normal lungs or pulmonary TB manifestations. Saliency visualizations generate heat-maps by measuring the derivative of the output class score concerning the original input. The resolution of saliency maps is higher compared to CAM-based visualizations. Figure 11 shows saliency map visualizations achieved with the VGG-16 fine-tuned model using an instance of abnormal CXR each from the Shenzhen TB CXR-Subset-1 and Montgomery TB CXR test set to visualize regions of TB manifestations.

Fig 11 (a) shows an instance of CXR from the Shenzhen TB CXR-Subset-1 dataset that is truly classified as showing TB manifestations. The CXR shows a patchy opacity in the right upper lung lobe. The saliency maps (Fig 11 (b)) show high activations on upper right lung lobe regions; these activations are consistent with the STAPLE-generated consensus annotation (Fig 11 (c)) that stands indicative of TB. However, there are a few FPs highlighted on the left upper lung lobe. These highlights are indicative of the highly sensitive nature of the model. This serves our goal to reduce potential FN radiologist errors before report generation. These regions will force the radiologist to verify their initial assessment and rule them out. We believe that such saliency visualization could help supplement clinical decisions while interpreting model predictions. Fig 11 (d) shows an instance of CXR from the Montgomery TB CXR dataset that is truly classified as having pulmonary TB. The saliency maps (Fig 11 (e)) highlight regions in the right upper lung lobe and these are consistent with the consensus annotation except that there is a faint insignificant activation observed above the right upper lung lobe (Fig 11 (f)).

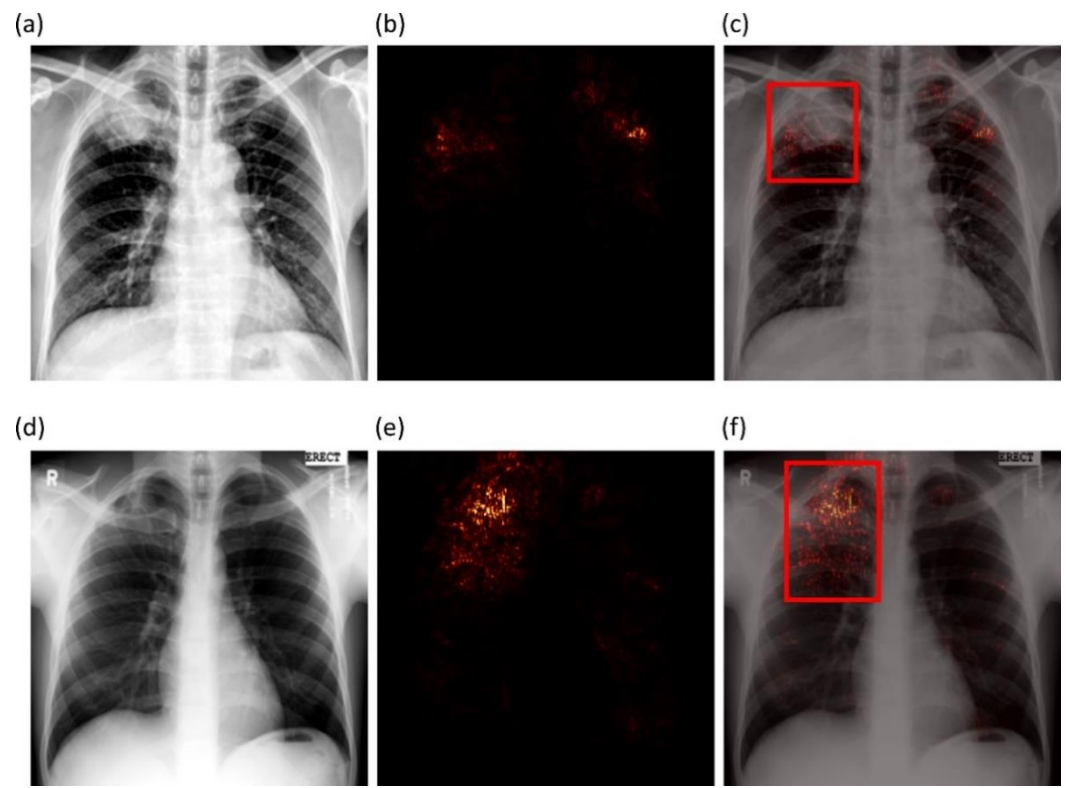

**Figure 11.** Saliency visualization. (**a**) and (**d**) An instance of Shenzhen TB CXR-Subset-1 and Montgomery TB CXR respectively; (**b**) and (**e**) Saliency activations; (**c**) and (**f**) Saliency overlaid on original images with expert GT annotation (bounding box in red) showing regions of TB manifestations most pronounced in the right apex.

We used the CRM-based visualization algorithm to compare the localization performance of baseline (ImageNet-pretrained DL models) and the fine-tuned models toward highlighting TB manifestations. Figure 12 shows the following: (i) an instance of Montgomery TB CXR with consensus annotation, and (ii) TB-consistent ROI localization achieved with various fine-tuned models and their baseline counterparts. Features are extracted from the deepest convolutional layer of the fine-tuned models and their baseline counterparts. CRM localization algorithm is then used to localize TB-consistent ROI pixels involved in the final predictions. It is observed from Figure 12 that the baseline models demonstrate sub-optimal TB-consistent ROI localization compared to the fine-tuned models. The TB-consistent ROI localization obtained using the fine-tuned models conform to the experts' knowledge of the problem under study. The feature map dimensions vary across the models. Hence, the CRMs are up-scaled through normalization methods to match the spatial resolution of the input image. The computed CRMs are overlaid on the original image to localize the TB-consistent ROI that is used to categorize the CXRs as showing pulmonary TB manifestations.

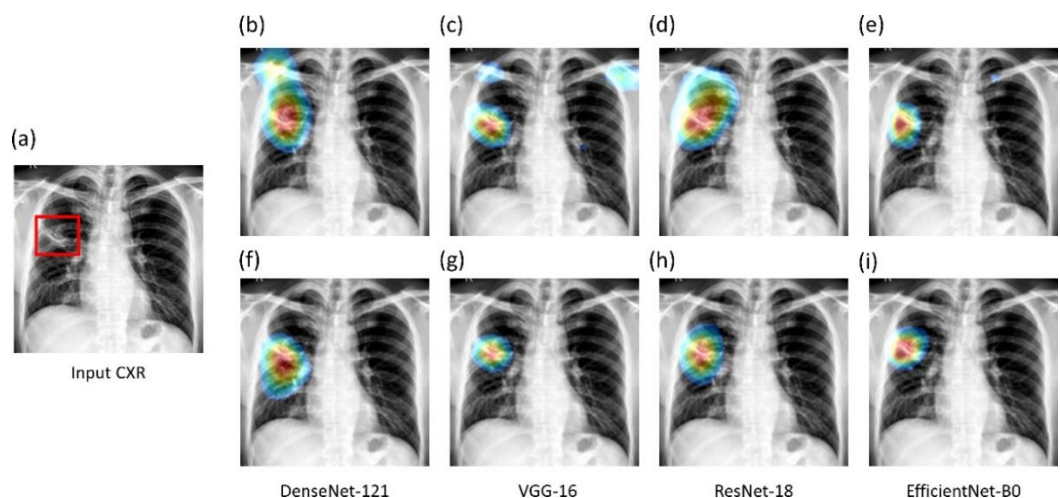

**Figure 12.** CRM-based TB-consistent ROI localization achieved using the CXR modality-specific pretrained/fine-tuned models and their baseline counterparts. (**a**) An instance of Montgomery TB CXR with STAPLE-generated consensus annotation (shown with a red bounding box); (**b**) and (**f**) baseline and fine-tuned DenseNet-121; (**c**) and (**g**) baseline and fine-tuned VGG-16; (**d**) and (**h**) baseline and fine-tuned ResNet-18; (**e**) and (**i**) baseline and fine-tuned EfficientNet-B0.

Figure 13 illustrates the sequence of steps involved in CRM-based TB-consistent ROI localization and binary mask generation.

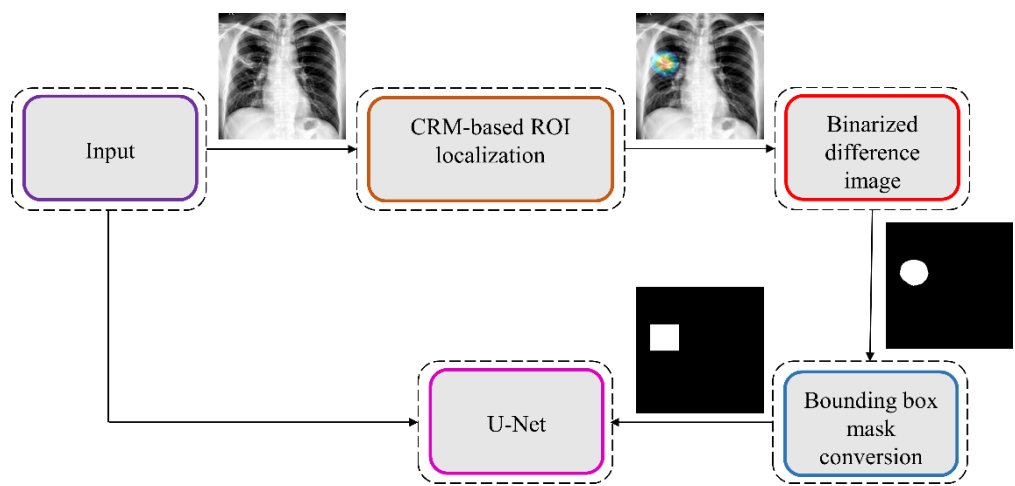

**Figure 13.** The process workflow showing the steps involved in CRM-based TB-consistent ROI localization and ROI mask generation.

The sequence of steps involved in this approach are mentioned as follows: (i) we computed the difference between the CRM-overlaid image and the original image and converted it into a binary image; (ii) The parameters of the polygonal coordinates of the connected components in the binary image are measured. This gives the coordinates of the vertices and that of the line segments making up the sides of the polygon; (iii) A binary mask is then generated from the polygon and stored; and (iv) The original images and their associated TB-consistent ROI binary masks are used for further analysis. It is observed from Table 5 that the models' performance with AT is superior compared to their baseline counterparts, i.e., without augmenting the training data. The proposed VGG16-CXR-U-Net (AT) demonstrated superior values for IOU and Dice metrics. However, concerning the AP@[.5:.95] metric that gives a balanced measure of precision and recall, the VGG19-U-Net (AT) delivered improved performance. This demonstrated that this model has better discrimination power compared to other models. However, except for the standard U-Net ($p < 0.05$), other U-Net models did not demonstrate statistically significantly superior performance for the AP@[.5:.95] metric (*p* > 0.05). Figure 14 illustrates the AUC-ROC curve and confusion matrix obtained by the VGG19-U-Net (AT) model toward this segmentation task. It is observed from the confusion matrix that the VGG19-U-Net (AT) model demonstrated lower FNs compared to FPs and hence higher sensitivity compared to precision toward segmenting TB-consistent manifestations using the TBX11K test set. This underscores the fact that fewer TB-consistent image pixels are classified as belonging to the background. Figure 15 shows a TBX11K test image on which the predicted TB-consistent ROI mask is overlaid to delineate regions showing TB manifestations.

*3.5. TB-consistent ROI Segmentation Evaluation Using TBX11K Train/Test Dataset*

Table 5 shows the segmentation performance achieved by the U-Net models with and without AT using the TBX11K train/test dataset.

**Table 5.** TB segmentation performance achieved using the U-Net models with and without AT.

| Model | Dice | IOU | AP@[.5:.95] | % Improvement in AP@[.5:.95] |
|---|---|---|---|---|
| Standard U-Net | 0.5335 | 0.3691 | 0.3361 (0.2706, 0.4016) | 5.02 |
| Standard U-Net (AT) | 0.6571 | 0.5004 | 0.3863 (0.3188, 0.4538) | |
| V-Net | 0.6913 | 0.5435 | 0.5084 (0.4391, 0.5777) | 2.22 |
| V-Net (AT) | 0.7186 | 0.5725 | 0.5306 (0.4614, 0.5998) | |
| Improved attention U-Net | 0.7032 | 0.5548 | 0.4605 (0.3914, 0.5296) | 2.06 |
| Improved attention U-Net (AT) | 0.7144 | 0.5656 | 0.4811 (0.4118, 0.5504) | |
| VGG16-U-Net | 0.7268 | 0.5812 | 0.5296 (0.4604, 0.5988) | 0.46 |
| VGG16-U-Net (AT) | 0.7304 | 0.5876 | 0.5342 (0.4650, 0.6034) | |
| VGG19-U-Net | 0.7417 | 0.5999 | 0.5328 (0.4636, 0.6020) | 2.96 |
| VGG19-U-Net (AT) | 0.7468 | 0.596 | **0.5624 (0.4936, 0.6312)** | |
| VGG16-CXR-U-Net | 0.7304 | 0.5854 | 0.528 (0.4588, 0.5972) | 1.64 |
| VGG16-CXR-U-Net (AT) | **0.7552** | **0.6168** | 0.5116 (0.4423, 0.5809) | |
| VGG19-CXR-U-Net | 0.7233 | 0.5768 | 0.4863 (0.4170, 0.5556) | 3.27 |
| VGG19-CXR-U-Net (AT) | 0.7424 | 0.6024 | 0.5190 (0.4497, 0.5883) | |

AT denotes augmenting the training data with the masks generated through weak TB-consistent ROI localization using the fine-tuned models and their associated original CXRs. Data in parenthesis are 95% CI for the AP@[.5:.95] values measured as the Binomial (Clopper-Pearson) "exact" method corresponding to separate 2-sided CI with individual coverage probabilities of √0.95. The performance of the standard U-Net model is significantly worse compared to other models (*p* < 0.05). The best performances are denoted by bold numerical values in the corresponding columns.

*3.6. TB-consistent ROI Segmentation Evaluation Using TBX11K Train Set and Cross-institutional Test Sets*

We further evaluated the segmentation performance of the U-Net models using the training data from the TBX11K dataset and individual cross-institutional Shenzhen TB CXR-Subset-2 and Montgomery TB CXR test sets. Here, we performed two sets of evaluations: (i) First, we trained the models using the TBX11K training data and tested the segmentation performance individually with cross-institutional Shenzhen TB CXR-Subset-2 and Montgomery TB CXR collections; (ii) Next, we augmented the training data of the TBX11K dataset with the ROI masks generated from weak TB-consistent ROI localizations and their associated original CXRs and evaluated the test performance individually with the Shenzhen TB CXR-Subset-2 and Montgomery TB CXR collections. Table 6 shows the segmentation performance achieved by the U-Net models, with and without AT, using the Shenzhen TB CXR-Subset-2 test set.

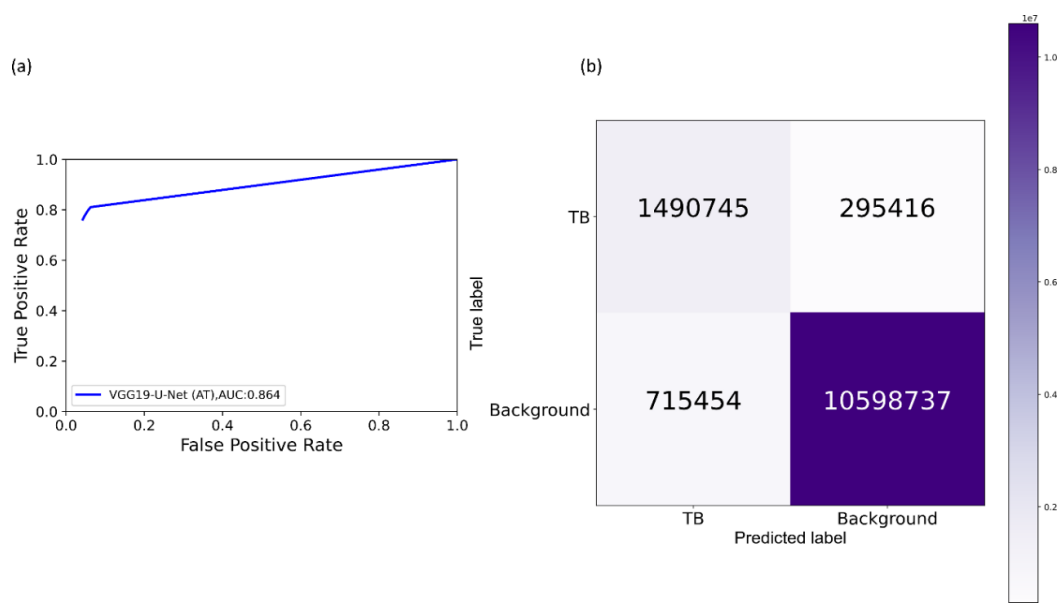

**Figure 14.** Performance achieved by the VGG19-U-Net (AT) toward TB-consistent ROI segmentation using TBX11K test set. (**a**) AUC-ROC curve; and (**b**) Confusion matrix.

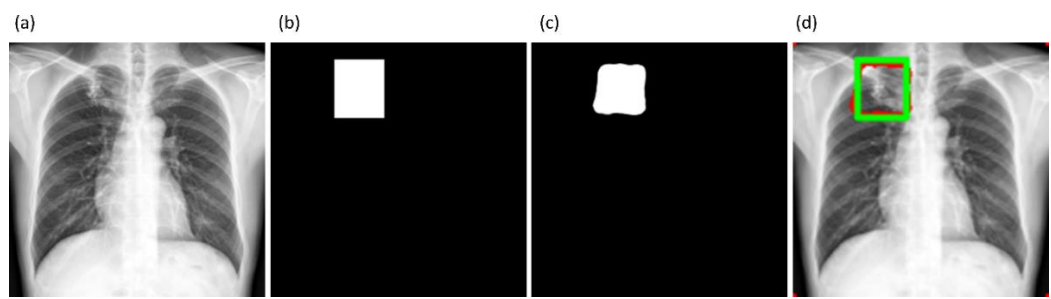

**Figure 15.** Sample overlay image. (**a**) An instance of TBX11K test CXR; (**b**) GT mask generated from bounding box annotations; (**c**) Mask predicted by the VGG19-U-Net (AT) model; and (**d**) GT and predicted mask overlaid on the original CXR (GT is denoted by a green bounding box and predicted mask is denoted by a red bounding box).

The performance achieved with the cross-institutional Shenzhen TB CXR-Subset-2 test set, as shown in Table 6, led to the following observations: (i) the performance of the models toward cross-institutional testing is not superior compared to using the test data coming from the TBX11K training distribution. This could be attributed to the changes in the visual characteristics of the images acquired across institutions. The Shenzhen TB CXR-Subset-2 test set is completely unseen during the training process; (ii) the models' performance with AT is superior compared to their baseline counterparts, i.e., without augmenting the training data; and (iii) the proposed VGG16-CXR-U-Net (AT) demonstrated statistically superior performance for IOU, Dice, and AP@[.5:.95] metrics compared to its baseline counterpart ($p < 0.05$). The model also performed statistically significantly better ($p < 0.05$) compared to other augmented/non-augmented trained models. Figure 16 illustrates the AUC-ROC curve and confusion matrix obtained by the proposed VGG16-CXR-U-Net (AT) model toward this segmentation task. Figure 17 shows an instance of Shenzhen TB CXR-Subset-2 test CXR on which the generated TB-consistent ROI mask is overlaid to delineate regions showing TB manifestations.

**Table 6.** TB-consistent ROI segmentation performance achieved by the U-Net models with and without AT using the Shenzhen TB CXR-Subset-2 test set.

| Model | Dice | IOU | AP@[.5:.95] | % Improvement in AP@[.5:.95] |
|---|---|---|---|---|
| Standard U-Net | 0.412 | 0.2639 | 0.3381 (0.2725,0.4037) | 1.66 |
| Standard U-Net (AT) | 0.5043 | 0.3429 | 0.3547 (0.2883,0.4211) | |
| V-Net | 0.4352 | 0.2856 | 0.4400 (0.3712,0.5088) | 4.72 |
| V-Net (AT) | 0.4836 | 0.3243 | 0.4872 (0.4179,0.5565) | |
| Improved attention U-Net | 0.4563 | 0.2997 | 0.4575 (0.3884,0.5266) | 6.76 |
| Improved attention U-Net (AT) | 0.4934 | 0.3344 | 0.5251 (0.4558,0.5944) | |
| VGG16-U-Net | 0.4754 | 0.3158 | 0.4506 (0.3816,0.5196) | 0.14 |
| VGG16-U-Net (AT) | 0.4951 | 0.3377 | 0.452 (0.383,0.521) | |
| VGG19-U-Net | 0.4692 | 0.3142 | 0.4849 (0.4156,0.5542) | 0.71 |
| VGG19-U-Net (AT) | 0.5114 | 0.3501 | 0.492 (0.4227,0.5613) | |
| VGG16-CXR-U-Net | 0.453 | 0.2979 | 0.4314 (0.3627,0.5001) | **10.3** |
| VGG16-CXR-U-Net (AT) | **0.5189** | **0.3503** | **0.5344 (0.4652,0.6036)** | |
| VGG19-CXR-U-Net | 0.4576 | 0.3025 | 0.5272 (0.458,0.5964) | 0.51 |
| VGG19-CXR-U-Net (AT) | 0.4694 | 0.3226 | 0.5323 (0.4631,0.6015) | |

Data in parenthesis are 95% CI for the AP@[.5:.95] values measured as the Binomial (Clopper-Pearson) "exact" method corresponding to separate 2-sided CI with individual coverage probabilities of √0.95. The best performances are denoted by bold numerical values in the corresponding columns.

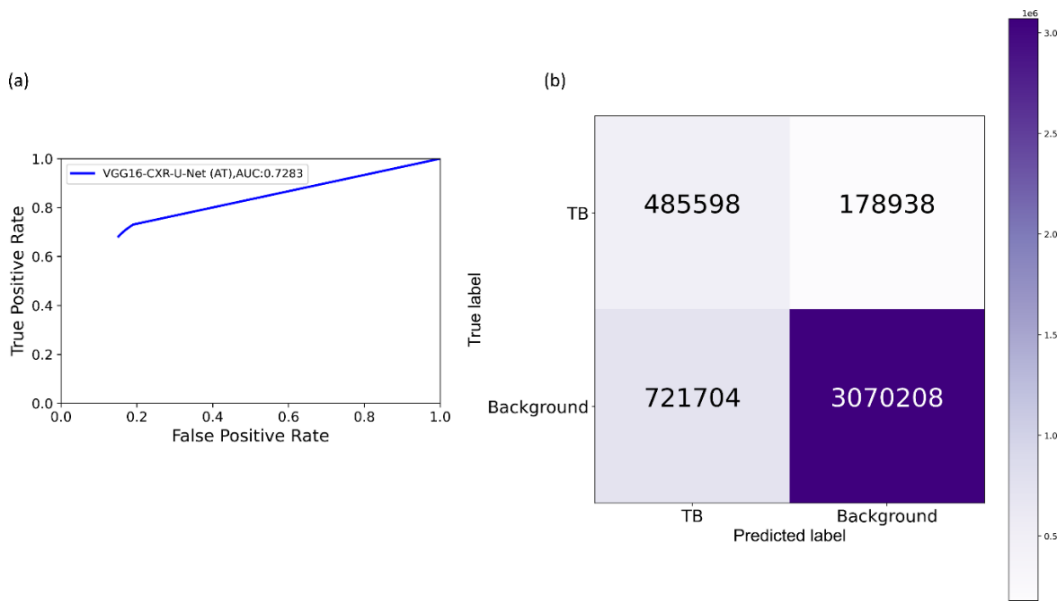

**Figure 16.** Performance achieved by the VGG16-CXR-U-Net (AT) toward TB-consistent ROI segmentation using Shenzhen TB CXR-Subset-2 test set. (**a**) AUC-ROC curve; and (**b**) Confusion matrix.

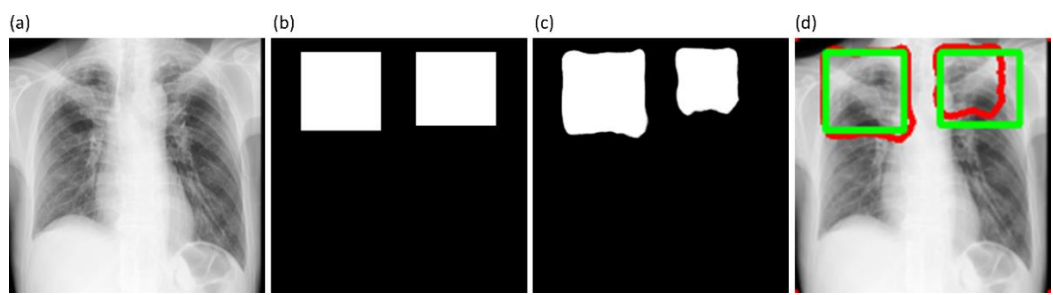

**Figure 17.** Sample overlay image. (**a**) An instance of Shenzhen TB CXR-Subset-2 test CXR; (**b**) GT mask generated from the STAPLE consensus ROI obtained using annotations of R2 and R3; (**c**) Predicted mask by the VGG19-U-Net (AT) model; and (**d**) GT and predicted mask overlaid on the original CXR (GT denoted by a green bounding box and predicted mask denoted by a red bounding box).

Table 7 shows the performance achieved through cross-institutional testing with the Montgomery TB CXR dataset.

**Table 7.** TB-consistent ROI segmentation performance achieved using the U-Net models with and without AT using the Montgomery TB CXR test set.

| Model | Dice | IOU | AP@[.5:.95] | % Improvement in AP@[.5:.95] |
|---|---|---|---|---|
| Standard U-Net | 0.301 | 0.1844 | 0.3171 (0.2526,0.3816) | |
| Standard U-Net (AT) | 0.4531 | 0.3021 | 0.3297 (0.2645,0.3949) | 1.26 |
| V-Net | 0.4121 | 0.2785 | 0.4114 (0.3432,0.4796) | |
| V-Net (AT) | 0.4365 | 0.3021 | 0.4682 (0.399,0.5374) | 5.68 |
| Improved attention U-Net | 0.3903 | 0.2573 | 0.4218 (0.3533,0.4903) | |
| Improved attention U-Net (AT) | 0.4806 | 0.3273 | 0.5418 (0.4727,0.6109) | 12.0 |
| VGG16-U-Net | 0.4263 | 0.2949 | 0.5275 (0.4583,0.5967) | |
| VGG16-U-Net (AT) | 0.4771 | 0.3211 | 0.539 (0.4699,0.6081) | 1.15 |
| VGG19-U-Net | 0.4238 | 0.2881 | 0.5029 (0.4336,0.5722) | |
| VGG19-U-Net (AT) | 0.4789 | 0.3148 | **0.6043 (0.5365,0.6721)** | 10.14 |
| VGG16-CXR-U-Net | 0.4667 | 0.3200 | 0.4191 (0.3507,0.4875) | |
| VGG16-CXR-U-Net (AT) | **0.5261** | **0.3743** | 0.5823 (0.5139,0.6507) | **16.32** |
| VGG19-CXR-U-Net | 0.4809 | 0.3306 | 0.4935 (0.4242,0.5628) | |
| VGG19-CXR-U-Net (AT) | 0.4694 | 0.3226 | 0.5272 (0.458,0.5964) | 3.37 |

Data in parenthesis are 95% CI for the AP@[.5:.95] values measured as the Binomial (Clopper-Pearson) "exact" method corresponding to separate 2-sided CI with individual coverage probabilities of $\sqrt{0.95}$. The best performances are denoted by bold numerical values in the corresponding columns.

Comparing these results with Table 5, we observed the following: (i) the performance of the models with cross-institutional Montgomery TB CXR test set is not superior compared to that using the test data coming from the TBX11K training distribution. These observations are analogous to those obtained with the Shenzhen TB CXR-Subset-2 test set, shown in Table 6. The inherent data variability across institutions resulted in sub-optimal model performance; (ii) the models' performance with AT using the Montgomery TB CXR test set is superior compared to non-augmented training; (iii) the proposed VGG16-CXR-U-Net (AT) demonstrated statistically superior values for IOU and Dice metrics ($p < 0.05$) compared to its baseline counterpart. However, concerning the AP@[.5:.95] metric, the VGG19-U-Net (AT) model demonstrated statistically significantly superior performance ($p < 0.05$) as compared to other models. Figure 18 illustrates the AUC-ROC curve and confusion matrix obtained using the VGG19-U-Net (AT) model toward this segmentation task. Figure 19 shows an instance of the Montgomery TB CXR test image on which the generated TB-consistent ROI mask is overlaid to delineate regions showing TB manifestations.

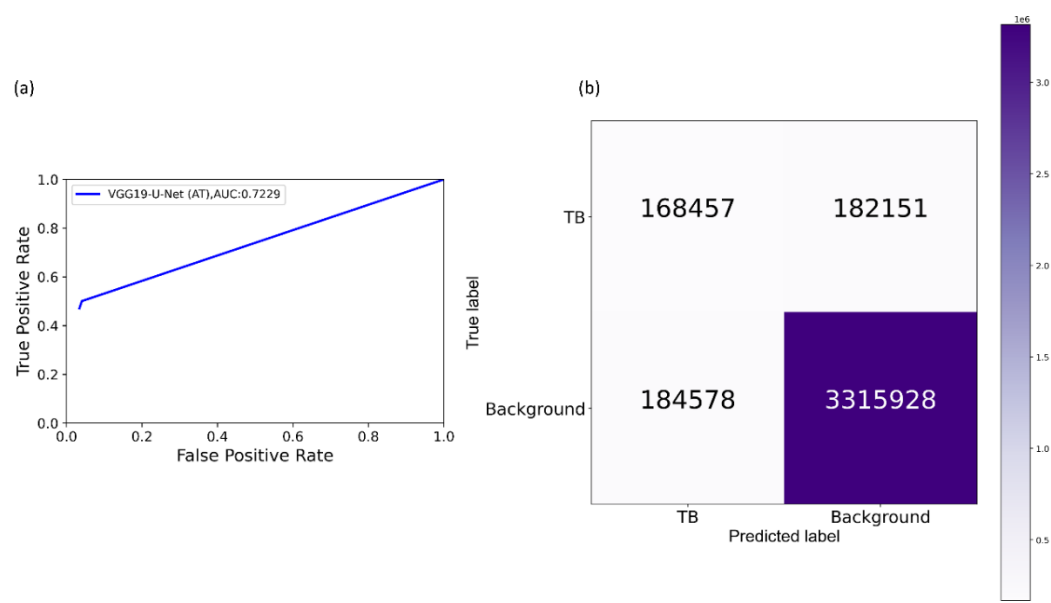

**Figure 18.** Performance achieved by the VGG19-U-Net (AT) toward TB-consistent ROI segmentation using the Montgomery TB CXR test set. (**a**) AUC-ROC curve; and (**b**) Confusion matrix.

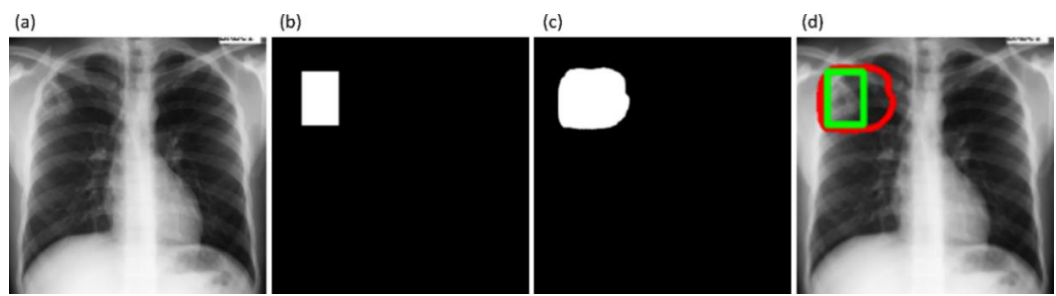

**Figure 19.** Sample overlay image. (**a**) An instance of Montgomery TB CXR; (**b**) GT mask generated from the STAPLE consensus ROI obtained using annotations of R1 and R2; (**c**) Mask predicted by the VGG19-U-Net (AT) model; and (**d**) GT and predicted mask overlaid on the original CXR (GT denoted by a green bounding box and predicted mask denoted by a red bounding box).

## 4. Discussion

We made several key observations from the results of this stage-wise systematic study. Broadly, these are the need for (i) modality-specific knowledge transfer techniques, (ii) verifying ROI localization, (iii) reducing inter-observer variability, (iv) selecting the appropriate loss function, (v) performing statistical analysis, (vi) evaluating with cross-institutional test collection, and (vii) selecting task-appropriate data augmentation methods. These observations are discussed in detail below.

Modality-specific knowledge transfer techniques: We note that CXR modality-specific pretraining helped improve performance in lung segmentation, fine-tuning, and TB segmentation and localization tasks over the non-modality-specific ImageNet-pretrained CNN models. Therefore, performing medical modality-specific training is a path toward improving model adaption and performance, reducing computational complexity and overfitting. The performance improvement may be attributed to the fact that the CXR modality-specific pretraining helped learn the characteristics of the lungs including shape, texture, contour, and their combinations that are diverse from natural stock photographic images. The learned modality-specific knowledge, when transferred to a relevant modality-specific task, helped in superior weight initialization compared to ImageNet-pretrained weights and resulted in improved performance.

The VGG-16 and VGG-19 CXR modality-specific segmentation models were better not only to localize but to classify the findings such as TB manifestations using the CXR modality-specific pretrained/fine-tuned classification models. This training approach facilitated a generalized modality-specific knowledge transfer that helped the segmentation models to better discriminate the lungs from the background pixels and the classification models to differentiate abnormal lungs with TB manifestations from normal lungs. An added benefit of this approach is that it resulted in a marked reduction in the number of trainable parameters.

Verifying ROI localization: Neural networks make use of inherent learning methods to identify salient features in the input data. However, data-driven DL models are considered back-boxes since they lack explainability. Therefore, it is crucial to determine if these models are predicting the expected classes for the right reasons. Saliency maps and other visual ablation studies help to investigate if the models learn salient feature representations conforming to the experts' knowledge of the problem. In this study, we used saliency maps and CRM-based localization algorithms to illustrate and explain the localization behavior of the trained models. We observed that the TB-consistent ROI localization obtained with the CXR modality-specific fine-tuned models are superior to those obtained using the ImageNet-pretrained models. Such superior localization performance could be attributed to the following characteristics: (i) the fine-tuned models inherit CXR modality-specific knowledge compared to that transferred from the natural image domain; (ii) the transferred knowledge is relevant to the target modality; (iii) the fine-tuned models are empirically truncated at the optimal depth to learn relevant and salient feature representations. This helped to deliver superior classification and localization performance. These observations are reinforced by the sub-optimal localization performance observed with the out-of-the-box ImageNet-pretrained models.

Reducing inter-observer variability: Literature studies are limited concerning arriving at a consensus ROI when using GT annotations from multiple experts. This is particularly important for medical decision-making since different experts may have varying opinions on the extent or location of the disease-specific ROI. The inherent bias in the experts' annotations may be due to a variety of explicit and implicit factors including overall clinical experience, specific background in the disease manifestations, and tendency to be aggressive or conservative when treating/detecting disease, or simply missing the disease leading to FNs in the data. To overcome these challenges, we used the STAPLE algorithm to arrive at a consensus ROI by discovering and quantifying the experts' bias and varying opinions about the TB-consistent ROI.

Selecting the appropriate loss function: Segmentation models often suffer from class imbalance issues, particularly in medical image segmentation tasks. This is due to the highly localized ROI, spanning for a very small percentage of the total number of image pixels. The issues with using the conventional BCE loss for such class imbalanced segmentation tasks are as follows: (i) the BCE loss weighs all image pixels equally; (ii) the model demonstrates low BCE loss and hence higher segmentation accuracy even if it misses all ROI pixels that spans a small portion of the total image. In this study, we used a customized loss based on Tversky Index, an asymmetric similarity measure that generalizes IOU and Dice metrics. The use of a customized Tversky Index-based loss function with empirically determined hyperparameter values for the problem under study helped to improve segmentation performance by providing a finer level of segmentation control compared to using a conventional BCE-based loss. Computing a single precision and recall score at a specific IOU threshold does not sufficiently describe the models' behavior. In this study, we used AP@[.5:.95] to effectively integrate the AUPRC by averaging the precision score at multiple IOU thresholds ranging from 0.5 to 0.95 (in 0.05 increments). Such a measure would help to better demonstrate models' generalization ability and stability toward the segmentation task.

Performing statistical analysis: Literature studies have shown that research publications seldom perform statistical significance analyses while interpreting their results [44]. We performed statistical analyses to investigate the existence of a statistically significant difference in the performance of the segmentation models based on the AP@[.5:.95] metric and classification models using the MCC metric. Such analysis helped to interpret model performance and other assumption violations.

Evaluating with cross-institutional test collection: It is indispensable to note that the data collected across institutions differ due to changes in imaging equipment, acquisition methods, processing protocols, and their combinations. These differences could notably affect their characteristics and render them qualitatively and quantitatively different across institutions. This may lead to sub-optimal performance when the models trained with the imaging data from one institution are tested with the data from another institution. The evaluation with cross-institutional test sets helps to empirically determine the existence of cross-institutional data variability and its effect on model robustness and generalization.

Selecting task-appropriate data augmentation technique: The performance of DL models is shown to improve with an increase in data and computational resources. Training data augmentation helps to significantly increase data diversity and thus the models' robustness and generalization to real-time applications. However, literature studies extensively discuss the use of affine-transformation-based data augmentation methods like cropping, flipping, rotating, and image padding. There is a need for progress in investigating the use of task-specific data augmentation strategies that could better capture inherent data variability, particularly toward medical image analyses, and help to improve performance. In this study, we observed that augmenting the training data with TB-specific weak localization improved TB segmentation and localization with both the test data coming from the same training distribution and cross-institutions. This performance improvement may be attributed to the fact that CXR modality-specific pretraining and task-specific training data augmentation, i.e., augmenting training data with weak, TB-consistent ROI localizations helped in added knowledge about CXRs and other diverse TB manifestations that resulted in superior weight initialization and improved segmentation performance with training distribution-similar and cross-institutional test sets.

Limitations: Regarding the limitations of the current study: (i) the numbers of publicly available CXRs showing pulmonary TB manifestations with expert ROI annotations are fairly small. In this study, we tried to alleviate this limitation by collecting expert annotations and generating STAPLE-based consensus ROI annotations for the publicly available Montgomery TB CXR and a subset of the Shenzhen TB CXR collections. However, future works could focus on training diversified models on large-scale CXR collections with sufficient data diversity and expert-annotated TB-specific ROIs and improve their confidence, robustness, and generalization toward real-time deployment. (ii) This study is evaluated to segment and localize TB-consistent findings. However, future research shall focus on testing against non-TB findings because TB-consistent findings including nodules and cavities are found in other abnormal pulmonary conditions. (iii) We augmented the training data with TB-specific weak ROI localizations to improve performance in a TB segmentation task. However, the effects of augmenting the training data with other pulmonary abnormality-specific localizations and the resulting performance are yet to be investigated. (iv) Considering limited data availability, this study proposes comparatively shallow VGG-16 and VGG-19 CXR modality-specific segmentation models be employed rather than other ImageNet-pretrained models toward the current task. With the availability of more annotated data, future research could propose diversified models with novel filters that may result in improved segmentation performance and reduction in computational complexity.

## 5. Conclusions

In this study we have demonstrated that CXR modality-specific pretraining/fine-tuning resulted in (i) transferring CXR modality-specific learned knowledge that can subsequently be fine-tuned to improve TB classification and segmentation performance; and (ii) improving segmentation performance with training distribution-similar and cross-institutional test sets by augmenting the training data with segmentation task-relevant, weak TB-consistent ROI localization. We have generated STAPLE consensus ROI from expert annotations for the publicly available Montgomery TB CXR and a subset of Shenzhen TB CXR collection that could be used by the research community (i) to train and evaluate segmentation and classification models and (ii) as a benchmark for developing effective computational methods. We believe that the results proposed in this study would be useful for developing robust models for classification, segmentation, and TB-consistent ROI localization tasks and eventually leading to more advanced assistance in radiologist interpretive workflows to include triage of abnormal findings and further classification based on distribution and other patterns.We believe this will eventually lead to improved patient care while improving productivity in addition to further providing quality labels via vetted annotations.

**Author Contributions:** Conceptualization, S.R. and S.A.; methodology, S.R. and S.A.; software, S.R.; validation, S.R. and S.A.; formal analysis, S.R. and S.A.; investigation, S.R. and S.A.; resources, S.A.; data curation, S.R., L.F., J.D., and P.A.; writing—original draft preparation, S.R. and S.A.; writing—review and editing, S.R., L.F., J.D., P.A. and S.A.; visualization, S.R.; supervision, S.A.; project administration, S.A.; funding acquisition, S.A. All authors have read and agreed to the published version of the manuscript.

**Funding:** This research was supported by the Intramural Research Program of the National Library of Medicine, National Institutes of Health.

**Data Availability Statement:** All data supporting the findings of this study are publicly available and are cited in the manuscript.

**Conflicts of Interest:** The authors declare no conflict of interest.